\documentclass[]{gensi}
\usepackage{tabularx}
\usepackage[toc,page,header]{appendix}

\usepackage{minitoc}
\usepackage{cleveref} 
\usepackage{subcaption}
\usepackage{booktabs}
\usepackage{graphicx}
\usepackage{pgfplots}
\usepackage{pgfplotstable}
\usepackage{xcolor}
\usepackage{CJKutf8}
\usetikzlibrary{patterns}
\usepackage{float}
\usepackage{caption}
\usepackage{algorithm}
\usepackage{algorithmic}
\usepackage{mdframed}
\usepackage{siunitx}

\usepackage{fix-cm}

\usepackage{natbib}
\usepackage{latexsym}

\usepackage{url}
\usepackage{amssymb}
\usepackage[utf8]{inputenc}
\usepackage{microtype}
\usepackage{booktabs}
\usepackage{pifont} 
\usepackage{multirow}
\usepackage{makecell}
\usepackage{paralist}
\usepackage{xspace}
\usepackage{color}
\usepackage{xcolor}
\usepackage{colortbl}
\usepackage{adjustbox}
\usepackage{hyperref} 
\usepackage[edges]{forest}
\usepackage{tikz} 
\usepackage{caption}
\usepackage{amsfonts}

\hypersetup{
    colorlinks,
    linkcolor={blue!80!black},
    citecolor={blue!80!black},
}
\tikzset{
    root/.style =             {align=center, text width=1cm, rounded corners=3pt, line width=0.3mm, fill=gray!10, draw=gray!80, font=\small},
    demographic/.style =         {align=center, text width=1.8cm, rounded corners=3pt, line width=0.3mm, fill=blue!10, draw=blue!80, font=\footnotesize},
    demographic_work/.style =    {align=center, text width=10cm, rounded corners=3pt, line width=0.3mm, fill=blue!10, draw=blue!0, font=\footnotesize},
    character/.style =         {align=center, text width=1.8cm, rounded corners=3pt, line width=0.3mm, fill=red!10, draw=red!80, font=\footnotesize},
    character_work/.style =    {align=center, text width=10cm, rounded corners=3pt, line width=0.3mm, fill=red!10, draw=red!0, font=\footnotesize},
    personalization/.style =           {align=center, text width=1.8cm, rounded corners=3pt, line width=0.3mm, fill=cyan!10, draw=cyan!80, font=\footnotesize},
    personalization_work/.style =      {align=center, text width=10cm, rounded corners=3pt, line width=0.3mm, fill=cyan!10, draw=cyan!0, font=\footnotesize},
    risk/.style =         {align=center, text width=1.8cm, rounded corners=3pt, line width=0.3mm, fill=orange!10, draw=orange!80, font=\footnotesize},
    risk_work/.style =    {align=center, text width=10cm, rounded corners=3pt, line width=0.3mm, fill=orange!10, draw=orange!0, font=\footnotesize},
}

\newenvironment{takeawaybox}
{\begin{mdframed}[backgroundcolor=gray!10, roundcorner=5pt, hidealllines=true, innerleftmargin=10pt, innerrightmargin=10pt, innertopmargin=8pt, innerbottommargin=8pt, skipabove=8pt, skipbelow=8pt]\noindent\textbf{Takeaway.}\ }
{\end{mdframed}}

\usepackage{CJK}

\title{
    Spectral Rewiring for Exploration, Purification, and Model Merging
}

\author[\ddagger, 1, 2, *]{Zhilong Zhang}
\author[1, 2, *]{Hongli Yu}
\author[1, 2]{Huan-ang Gao}
\author[1, 2]{Hanlin Wu}
\author[1, 2]{Yuxuan Song}
\author[1, 2]{Wei-Ying Ma}
\author[1, 2]{Ya-Qin Zhang}
\author[\dagger, 1, 2]{Hao Zhou}

\affiliation[1]{SIA-Lab of Tsinghua AIR and ByteDance Seed}
\affiliation[2]{Institute for AI Industry Research (AIR), Tsinghua University}

\contribution[*]{Equal Contribution}
\contribution[\dagger]{Corresponding Authors}
\contribution[\ddagger]{Project Lead}
\abstract{
    Reinforcement learning has become a standard post-training recipe for large language models, but dense full-parameter updates create two deployment-relevant bottlenecks: suppressed reasoning performance, often reflected by premature saturation of test-time scaling, and interference when consolidating multiple capabilities through multi-domain training or model merging. We show that the reasoning-effective component of these updates is largely concentrated in the base model's spectral space, motivating Subspace-Aligned Rewiring (SAR), a post-hoc editing method that retains this spectral core while removing orthogonal components. SAR therefore preserves reasoning gains and filters residual update directions that suppress performance or amplify cross-domain interference. Across several model families and scales, SAR \textbf{extracts} compact reasoning cores using as little as $\sim$0.58\% of total parameters: it preserves over 99\% of post-training performance and improves high-$k$ exploration in mathematical reasoning, and generalizes to agentic coding by improving six of seven open benchmarks on an in-house model. SAR also \textbf{purifies} mixed-domain training updates by releasing suppressed coding capability while maintaining math reasoning and instruction following. It further enables \textbf{model merging} across experts, yielding cross-domain generalization that surpasses previous merging baselines and even the best single-domain experts. Overall, SAR shows that extracting reasoning-effective updates from parameter geometry can serve as a training-free mechanism to improve reasoning and multi-domain performance.

}

\begin{document}

    \maketitle

    \footnotetext{Correspondence: zhouhao@air.tsinghua.edu.cn}

    \section{Introduction}

Reinforcement learning has become a central post-training paradigm for improving large language models. By optimizing models against task-level feedback--such as mathematical answers, code execution results, and success signals in agentic workflows--recent systems can improve capabilities in mathematics, coding, and broader interactive settings \cite{shao2024deepseekmath,guo2025deepseekr1,cobbe2021training,lightman2023lets}.

Despite its empirical success, the ``black-box'' nature of full-parameter RL post-training offers limited control over the induced parameter update, creating two practical bottlenecks for large-scale reasoning models. \textit{First}, it can induce \textbf{suppressed reasoning performance}, often visible as saturation of test-time scaling: reward optimization may concentrate the policy around a narrow set of high-reward trajectories, reducing the diversity of candidate solutions and limiting the ability to solve harder instances \cite{kirk2023understanding,moskovitz2023confronting,brown2024large,snell2024scaling}. \textit{Second}, it can induce \textbf{cross-domain interference} during capability consolidation: updates from multiple domains, such as mathematics, competitive programming, and instruction following, may occupy incompatible directions in parameter space, so joint RL or weight-space merging can improve one capability while suppressing another \cite{ilharco2022editing,yadav2023ties,yu2023language}. This motivates a key question: can we identify a coordinate system that separates reasoning-effective update components from residual directions, thereby recovering suppressed reasoning performance and improving multi-domain generalization?

\begin{figure}[!t]
    \vspace{-7.0em}
    \centering
    \includegraphics[width=0.9\linewidth]{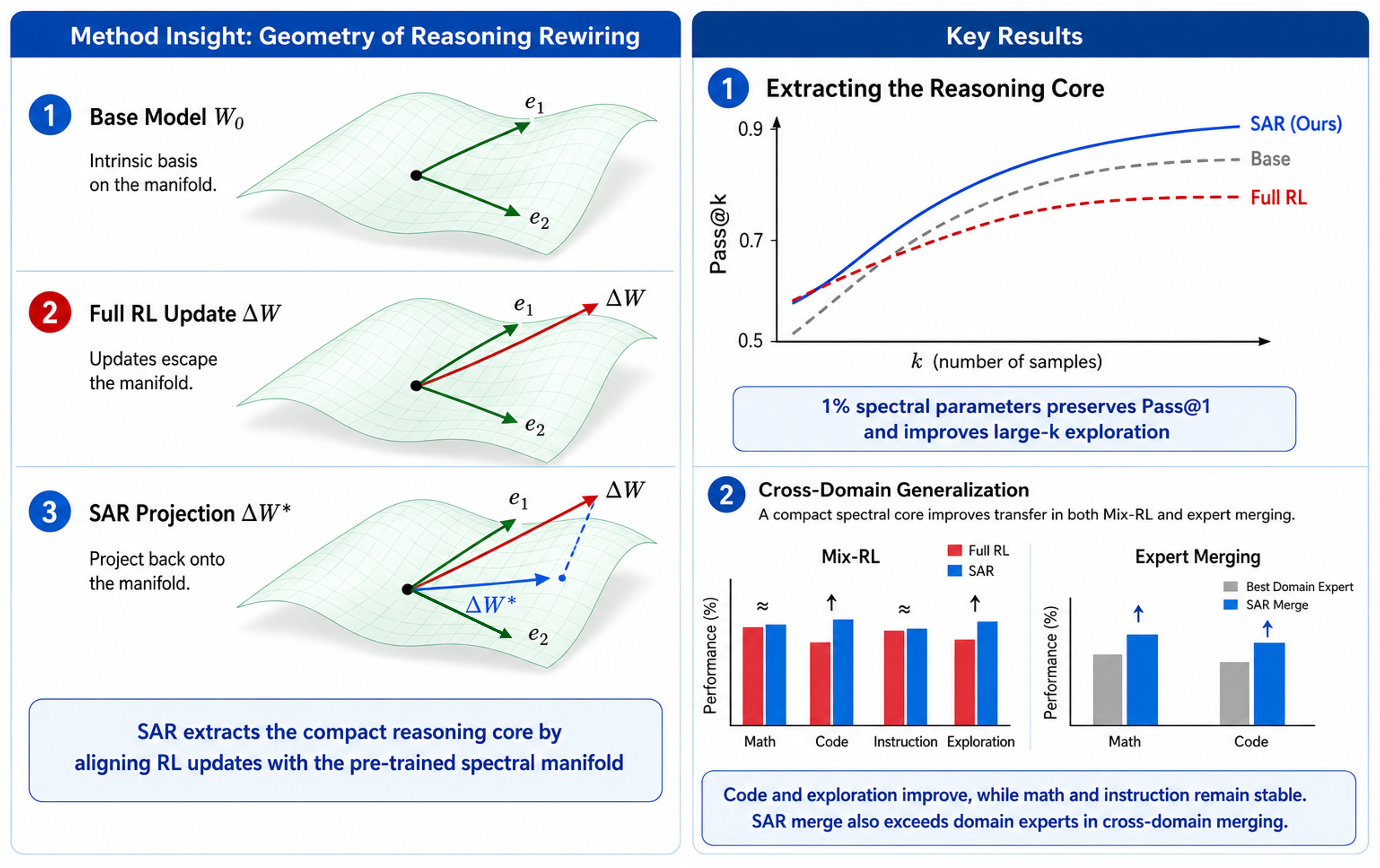}
    \caption{Overview of Subspace-Aligned Rewiring (SAR). SAR identifies a compact reasoning core by projecting reinforcement learning updates back onto the pretrained model's spectral manifold. This alignment preserves the intrinsic reasoning basis of the base model while removing off-manifold update components introduced by full RL. The resulting spectral update retains Pass@1 performance, improves large-$k$ exploration, and transfers more robustly across domains, including Mix-RL and expert merging.}
    \label{fig:placeholder}
\end{figure}

We propose a coordinate system for studying reasoning updates from a geometric perspective. Outcome-reward RL often elicits knowledge and skills that already exist in the base model, so the base model and the RL model should remain \textbf{closely related in an appropriate geometric manifold}. We propose that this manifold is given by the base model's SVD coordinates. In this view, the singular vectors of the base weights represent pretrained ``atomic skills,'' while the diagonal singular-value matrix specifies how these skills are originally connected. RL keeps this pretrained basis largely fixed, but changes the middle matrix from a diagonal map into a non-diagonal \textbf{rewiring matrix}. This rewiring reconnects atomic skills that were previously isolated, allowing the model to compose existing knowledge into new reasoning circuits.

This perspective leads to Subspace-Aligned Rewiring (SAR), a post-hoc model editing method that projects a dense reasoning update onto the pretrained spectral manifold. The resulting spectral update supports three application-facing uses. \textbf{Extraction}: SAR isolates a compact reasoning-effective component that preserves post-training performance across 1.5B--32B public models, improves high-$k$ exploration, and improves open agentic coding benchmarks on an in-house model. \textbf{Purification}: SAR filters residual directions in Mix-RL updates, reducing domain conflict and releasing suppressed cross-domain capability. \textbf{Merging}: SAR places expert updates in a common pretrained coordinate system, improving multi-domain model merging beyond previous baselines and even the best single-domain experts. Our contributions are organized as four takeaways:

\begin{itemize}
    \item \textbf{Compact Spectral Rewiring Extracts Effective Post-Training Signal.}
    Across 1.5B--32B reasoning models, SAR preserves full-RL-level math performance with as little as \textbf{$\sim$0.58\%} of total parameters and \textbf{more than 99\% of peak $Pass@1$ preserved}. Beyond math, the same top-1\% spectral projection improves six of seven open agentic coding benchmarks on an in-house model, with a \textbf{+2.52\% improvement on average}.

    \item \textbf{Spectral Projection Restores Useful Exploration.}
    SAR improves exploration beyond the base model and \textbf{surpasses unconstrained full-parameter RL baselines in Pass@$k$ scaling}, crossing the full-RL curve at small $k$ and expanding the set of AIME problems reachable through repeated sampling.

    \item \textbf{Spectral Projection Improves Mix-RL.}
    In Mix-RL, SAR reveals \textbf{suppressed cross-domain capability}. On OLMo-3.1-32B-Think, SAR improves competitive coding (+1.48\% on LiveCodeBench v5 and +0.95\% on v6), improves mathematical exploration (AIME 25 $Pass@16$ from 88.33\% to 91.71\%), and maintains strong instruction-following performance.

    \item \textbf{Expert Merging Surpasses Domain Experts.}
    In math-code expert merging, SAR produces merged models that \textbf{surpass the best single-domain experts} at both 1.5B and 14B scales, improving AIME and LiveCodeBench simultaneously.
\end{itemize}


\section{The Geometric Framework of Reasoning Elicitation}
\label{sec:geometric_framework}

In this section, we construct a first-principles geometric framework to formalize how outcome-reward RL elicits reasoning capabilities from a pretrained model. We first establish the spectral subspace of the base model as a functional basis for reasoning. We then introduce Subspace-Aligned Rewiring (SAR), a projection-based algorithm designed to extract reasoning-effective updates geometrically. Finally, we provide a mechanistic analysis of how reasoning can emerge through spectral rewiring.

\subsection{The Functional Basis: SVD in Forward Propagation}
\label{subsec:functional_basis}
Consider a linear layer within a pretrained base model, parameterized by a weight matrix $W_0 \in \mathbb{R}^{d_{out} \times d_{in}}$. The forward propagation for an input representation $x$ is defined as $y_{base} = W_0 x$. To geometrically dissect this transformation, we apply singular value decomposition (SVD) to the weight matrix:
\begin{equation}
\label{eqs:base_forward}
y_{base} = W_0x = U \Sigma V^\top x = \sum_{i=1}^r \sigma_i u_i (v_i^\top x)
\end{equation}
where $r$ denotes the rank, while $V = [v_1, \dots, v_r]$ and $U = [u_1, \dots, u_r]$ represent the right and left singular vectors of $W_0$, respectively. 

As shown in Equation~\ref{eqs:base_forward}, the transformation decomposes into two phases: (1) \textbf{Information Read-in}: the right singular vectors $v_i$ act as principal feature detectors, extracting activation scores $(v_i^\top x)$ from the input; and (2) \textbf{Information Read-out}: the left singular vectors $u_i$ define the principal output directions, weighted by the singular values $\sigma_i$ and $(v_i^\top x)$.

Based on this analysis, these singular vectors constitute the \textbf{latent skills} inherent in the base model, functioning as the geometric basis to \textbf{encode input signals} and \textbf{decode output representations}. Furthermore, extensive literature in model compression and spectral analysis \cite{han2015deep,yuan2023asvd,wang2024svdllm} suggests that the vast majority of a pretrained model's functional capacity is bottlenecked within this principal spectral subspace. Consequently, we define the \textbf{pretrained spectral manifold} of a matrix space spanned by this coordinate system:
\begin{equation}
\label{eq:spectral_manifold}
    \mathcal{S}_r(W_0) = \text{span}(U \otimes V),
\end{equation}
where $U_r,V_r$ are the singular vectors of $W_0$ with $r = \min(d_{out},d_{in})$.
This subspace provides a compact coordinate system for the model's functional library. In the following section, we leverage this geometric foundation to introduce our core insights into the mechanisms of reasoning elicitation.

\subsection{The Geometric Principle of Reasoning and Spectral Projection}
\label{subsec:hypothesis_and_algorithm}
We propose a \textbf{first-principles geometric hypothesis} for reasoning elicited by outcome-reward RL: \textit{\textbf{the reasoning-effective component of a full reasoning update should be recoverable within the pre-existing spectral subspace of the base model.}}

Let $W_{RL}$ denote a model obtained by full-parameter outcome-reward RL and define the empirical update
\begin{equation}
    \Delta W = W_{RL} - W_0.
\end{equation}

The update $\Delta W $ is trained without geometric constraints, so it can be orthogonally decomposed into two components:
\begin{equation}
\label{eq:decompose_update}
    \Delta W = \Delta W^* + \Delta W_\perp,
\end{equation}
where $\Delta W^* \in \mathcal{S}_k(W_0)$ is the subspace-aligned update associated with measured reasoning gains, and $\Delta W_\perp$ is the residual outside the
pretrained spectral manifold. 

Using the projection operators $P_U = UU^\top$ and $P_V = VV^\top$, we formalize the extraction of the reasoning component as:
\begin{equation}
    \begin{aligned}
    \Delta W^* &= P_U \Delta W P_V \\
    &= (U U^\top) \Delta W  (V V^\top)\\
    &= U (U^\top \Delta W  V) V^\top \\
    &= U M V^\top
    \end{aligned}
    \label{eq:sar_projection}
\end{equation}
where 
$$M  = U^\top \Delta W  V\in \mathbb{R}^{r \times r}$$ 

is the \textbf{Rewiring Matrix}. Unlike the diagonal matrix $\Sigma$, $M$ is generically dense, capturing the \textbf{cross-dimensional interactions} between pretrained singular vectors. 

This process, summarized in Algorithm~\ref{alg:sar}, extracts a compact reasoning-effective representation $\Delta W^*$ while discarding the residual component $\Delta W_\perp$. In practice, we first extract the \textbf{top-$k$ low-rank component of $\Delta W$} and then project it onto the pretrained SVD subspace, thereby combining parameter compression with spectral alignment to identify the compact geometry responsible for reasoning. Appendix~\ref{app:evaluation_config} specifies the exact parameter groups, precision settings, and evaluation configuration used in our implementation.

\begin{algorithm}[H]
\caption{Subspace-Aligned Rewiring (SAR)}
\label{alg:sar}
\begin{algorithmic}[1]
\renewcommand{\algorithmicrequire}{\textbf{Input:}}
\renewcommand{\algorithmicensure}{\textbf{Output:}}
\REQUIRE Base model $W_0$, RL model $W_{RL}$, target rank $k$
\ENSURE Projected reasoning model $W_{SAR}$
\STATE \textbf{Spectral Extraction:} Compute top-$k$ SVD of $W_0 = U \Sigma V^\top$
\STATE \textbf{Isolate Update:} $\Delta W \leftarrow W_{RL} - W_0$
\STATE \textbf{Low-Rank Extraction:} Extract the top-$k$ component $\Delta W_k$ from $\Delta W$
\STATE \textbf{Geometric Projection:} Extract rewiring matrix $M \leftarrow U^\top \Delta W_k V$
\STATE \textbf{Reconstruction:} $\Delta W^* \leftarrow U M V^\top$ \COMMENT{Aligns the compact update with the pretrained spectral subspace}
\RETURN $W_{SAR} \leftarrow W_0 + \Delta W^*$
\end{algorithmic}
\end{algorithm}

\subsection{Mechanistic View: Reasoning as Rewiring}
\label{subsec:mechanistic_proof}

To analyze how the extracted rewiring matrix $M$ can capture complex reasoning behavior, we examine the forward propagation of the SAR-projected model: 
$$W^* = W_0+ \Delta W^* = U (\Sigma + M) V^\top$$
For an input $x$, the rewired forward pass is:
\begin{equation}
\label{eq:rewired_forward}
\begin{aligned}
    y_{rewired} &= U (\Sigma + M) V^\top x \\
    &=\sum_{i=1}^r \left[ \underbrace{(\sigma_i + M_{ii}) (v_i^\top x)}_{\text{Rescaling}} + \underbrace{\sum_{j \neq i} M_{i\textcolor{red}{j}} (v_{\textcolor{red}{j}}^\top x)}_{\text{Rewiring}} \right] u_i
\end{aligned}
\end{equation}
Equation~\ref{eq:rewired_forward} reveals the dual nature of RL updates within the spectral subspace:
\begin{itemize}
\item \textbf{Diagonal Elements ($M_{ii}$):} These represent the rescaling of isolated, pre-existing capabilities. While they amplify certain skills, they do not establish new logical connections.
\item \textbf{Off-Diagonal Elements ($M_{ij}$):} These provide a geometric mechanism for reasoning. They enable an output direction $u_i$ to be synthesized by \textbf{multiple, distinct input features $v_j$ ($j \neq i$)}. This transition from \textbf{one-to-one associative mapping to many-to-one logical synthesis} allows the model to perform multi-conditional deductions.
\end{itemize}

The algebraic distinction between diagonal and off-diagonal dynamics in $M$ provides a structural analogy to relational reasoning \cite{halford1998processing,santoro2017simple,battaglia2018relational}. Diagonal scaling $M_{ii}$ resembles unary associative memory: a single latent cue $v_i^\top x$ activates an isolated output concept along $u_i$. Off-diagonal entries $M_{ij}$, by contrast, instantiate cross-component integration, routing evidence from one latent premise direction $v_j$ toward a different conclusion direction $u_i$. When multiple such terms are active, the network no longer performs simple point-to-point retrieval; it \textbf{composes independent latent premises into a shared output representation}. This offers a mechanistic interpretation of why off-diagonal spectral rewiring can support reasoning elicitation: $\Delta W^\ast$ preserves the geometric structure associated with relational integration, while removing update components outside this spectral computation. Appendix~\ref{app:mechanistic_example} provides an illustrative example.

\begin{takeawaybox}
\begin{itemize}
    \item The pretrained SVD basis provides a functional coordinate system for reasoning updates, turning a dense weight change into an editable spectral object.
    \item In this coordinate system, post-training can be viewed as changing how pretrained skill directions are connected, rather than replacing the model's learned basis.
    \item This perspective gives a practical interface for downstream operations: compression, filtering, and merging can be applied to the rewiring matrix instead of the full parameter update.
\end{itemize}
\end{takeawaybox}

In the subsequent sections, we empirically validate these theoretical insights through extensive experiments, demonstrating that SAR preserves peak reasoning performance, improves useful exploration and open agentic coding performance on an in-house model, and strengthens cross-domain generalization.

\section{Extracting the Reasoning Core: Single-Domain Validation}
\label{sec:accuracy_diversity}

In this section, we demonstrate that SAR can extract a compact spectral update that preserves full post-training performance on mathematical reasoning tasks, improves high-$k$ exploration, and boosts open agentic coding benchmark performance on an in-house model. Together, these results show that extraction is not merely a compression result, but a practical post-hoc editing operation for improving downstream utility.

\subsection{Spectral Rewiring Preserves Reasoning Gains}
\label{subsec:geometric_sufficiency}

\begin{table*}[t]
\centering
\renewcommand{\arraystretch}{1.35}
\caption{Reasoning performance on AIME 2024 and AIME 2025 across different model scales and training recipes. SAR reports the smallest retained spectral rank found by rank sweep that preserves full-RL-level performance within evaluation variance.}
\label{tab:aime_results}
\resizebox{\textwidth}{!}{
\begin{tabular}{llcccc}
\toprule
\multirow{2}{*}{\textbf{Model Size}} & \multirow{2}{*}{\textbf{Method}} & \multicolumn{2}{c}{\textbf{AIME 24}} & \multicolumn{2}{c}{\textbf{AIME 25}} \\
\cmidrule(lr){3-4} \cmidrule(lr){5-6}
& & \textbf{AVG@32} & \textbf{Pass@32} & \textbf{AVG@32} & \textbf{Pass@32} \\
\midrule

\multicolumn{6}{l}{\cellcolor{gray!20}The SOTA Training Recipes} \\
\midrule
& Base: Deepseek-distill-qwen2.5 & 31.67\% & 80.00\% & 23.96\% & 53.33\% \\
\cmidrule(lr){2-6}
& RL: DeepscaleR & \textbf{40.31\%} & \textbf{76.67\%} & \textbf{29.58\%} & \textbf{60.00\%} \\
\cmidrule(lr){2-6}
\multirow{-3}{*}{\textbf{1.5B}} 
& \cellcolor{gray!10}\textbf{Projected (Ours): 1\%}  & \cellcolor{gray!10}40.21\% & \cellcolor{gray!10}\textbf{76.67\%} & \cellcolor{gray!10}28.96\% & \cellcolor{gray!10}\textbf{60.00\%} \\
\midrule

& Base: Qwen3 4B & 72.50\% & 90.00\% & 63.44\% & 86.67\% \\
\cmidrule(lr){2-6}
& RL: Polaris & \textbf{78.75\%} & \textbf{93.33\%} & \textbf{75.21\%} & 90.00\% \\
\cmidrule(lr){2-6}
\multirow{-3}{*}{\textbf{4B}} 
& \cellcolor{gray!10}\textbf{Projected (Ours): 10\%} & \cellcolor{gray!10}78.02\% & \cellcolor{gray!10}\textbf{93.33\%} & \cellcolor{gray!10}72.21\% & \cellcolor{gray!10}\textbf{93.33\%} \\
\midrule

& Base: OLMo3-32B-think-dpo & 74.48\% & 93.33\% & 70.83\% & 90.00\% \\
\cmidrule(lr){2-6}
& RL: OLMo-3.1-32B-Think & \textbf{79.56\%} & \textbf{93.33\%} & \textbf{75.76\%} & 90.00\% \\
\cmidrule(lr){2-6}
\multirow{-3}{*}{\textbf{32B}} 
& \cellcolor{gray!10}\textbf{Projected (Ours): 1\%} & \cellcolor{gray!10}78.88\% & \cellcolor{gray!10}\textbf{93.33\%} & \cellcolor{gray!10}75.12\% & \cellcolor{gray!10}\textbf{93.33\%} \\
\midrule

\multicolumn{6}{l}{\cellcolor{gray!20}Emergent Reasoning Ability from the Base Model} \\
\midrule
& Base: OLMo3-7B-Base & 22.08\% & 70.00\% & 19.58\% & 60.00\% \\
\cmidrule(lr){2-6}
& RL: OLMo-3.1-7B-RL-Zero-Math & \textbf{50.21\%} & 76.67\% & \textbf{36.88\%} & \textbf{76.67\%} \\
\cmidrule(lr){2-6}
\multirow{-3}{*}{\textbf{7B}} 
& \cellcolor{gray!10}\textbf{Projected (Ours): 30\%} & \cellcolor{gray!10}49.02\% & \cellcolor{gray!10}\textbf{80.00\%} & \cellcolor{gray!10}36.56\% & \cellcolor{gray!10}\textbf{76.67\%} \\

\bottomrule
\end{tabular}
}
\end{table*}

To assess the compactness of spectral rewiring, we apply SAR to extract the reasoning-effective component $\Delta W^*$ from open-source RL models and compare the projected models with their full RL counterparts. We perform a rank sweep and report the smallest retained spectral rank that preserves full-RL performance within evaluation variance. Appendix~\ref{app:no_projection_control} and Appendix~\ref{app:method_ablations} further compare SAR with no-projection and alternative projection controls.

\paragraph{Experimental Setup.}
We evaluate a diverse set of model families and scales, ranging from 1.5B to 32B parameters. Specifically, we consider DeepScaleR (1.5B) \cite{luo2025deepscaler}, POLARIS (4B) \cite{polaris2025preview}, and OLMo-3.1-32B-Think \cite{allenai2025olmo3} as representative strong RL reasoning models. We also evaluate OLMo-3-7B-Base and its RL-trained counterpart to study reasoning elicitation from a base model not explicitly specialized for mathematical reasoning \cite{allenai2025olmo3}. We use AIME 2024 and AIME 2025 as challenging mathematical reasoning benchmarks, following common practice in recent reasoning-model evaluations \cite{guo2025deepseekr1,luo2025deepscaler,hu2025openreasonerzero}. For each prompt, we sample 32 responses and report AVG@32 and Pass@32 \cite{chen2021evaluating}. Additional evaluation details are provided in Appendix~\ref{app:evaluation_config}.

\paragraph{Results.} As detailed in Table~\ref{tab:aime_results}, SAR preserves nearly all of the measured reasoning gains obtained by full RL across model families and scales. Across AIME 2024 and AIME 2025, the projected models match their corresponding full RL models within a small margin under AVG@32, while often matching or slightly improving Pass@32. These results suggest that the evaluated RL-induced reasoning improvements are \textbf{largely concentrated in the pretrained spectral subspace}, rather than requiring the full unconstrained parameter update. The required rank also reflects the role of the starting model: strong post-trained or distilled reasoning recipes require only a small retained rank, whereas eliciting reasoning directly from OLMo3-7B-Base requires a larger 30\% rank, suggesting that RL must modify a broader portion of the pretrained spectral coordinates when the starting model is less specialized for reasoning.

\paragraph{The Density of Reasoning-Effective Parameters.} The rank sweep identifies a small saturation point for the strong reasoning recipes: retaining only 1\% to 10\% of the pretrained spectral rank recovers full-RL-level reasoning performance for the evaluated models. As shown in Table~\ref{tab:parameter_efficiency}, the corresponding rewiring matrix $M$ is much smaller than both the full model and the reconstructed update $\Delta W^*$, with storage ratios as low as $\sim$0.58\% and $\sim$0.64\% of total parameters for the 1.5B and 32B models, respectively. This indicates that the reasoning-effective part of the RL update is highly concentrated in a compact spectral representation. Rather than suggesting that all discarded parameters are useless in general, these results show that a large fraction of the unconstrained RL update is unnecessary for preserving the measured AIME reasoning gains.

\begin{table}[htbp]
\centering
\caption{Parameter efficiency and compression ratios of SAR across different model scales. The intrinsic rewiring matrix $M$ is highly compact relative to the total parameter count.}
\label{tab:parameter_efficiency}
\resizebox{\textwidth}{!}{%
\begin{tabular}{lcccccc}
\toprule
\multirow{2}{*}{\textbf{Model Name}} & \textbf{Total} & \textbf{Projected} & \textbf{Parameter Update} & \textbf{Rewiring Matrix} & \textbf{Compression Ratio} \\
& \textbf{Params} & \textbf{Rank} & \textbf{($\Delta W^*$)} & \textbf{($M$)} & \textbf{($M$ / Total)} \\
\midrule
DeepScaleR & 1.54B & 1\% & 16.1M & 9.0M & $\sim$0.58\% \\
POLARIS & 4.00B & 10\% & 552.0M & 310.0M & $\sim$7.75\% \\
OLMo-3.1-32B-Think & 32.00B & 1\% & 311.0M & 204.0M & $\sim$0.64\% \\
\bottomrule
\end{tabular}%
}
\end{table}

\subsection{Spectral Rewiring Improves High-\texorpdfstring{$k$}{k} Exploration}
\label{subsec:curing_mode_collapse}

\begin{figure}[htbp]
     \centering
     \begin{subfigure}[b]{0.48\textwidth}
         \centering
         \includegraphics[width=\linewidth]{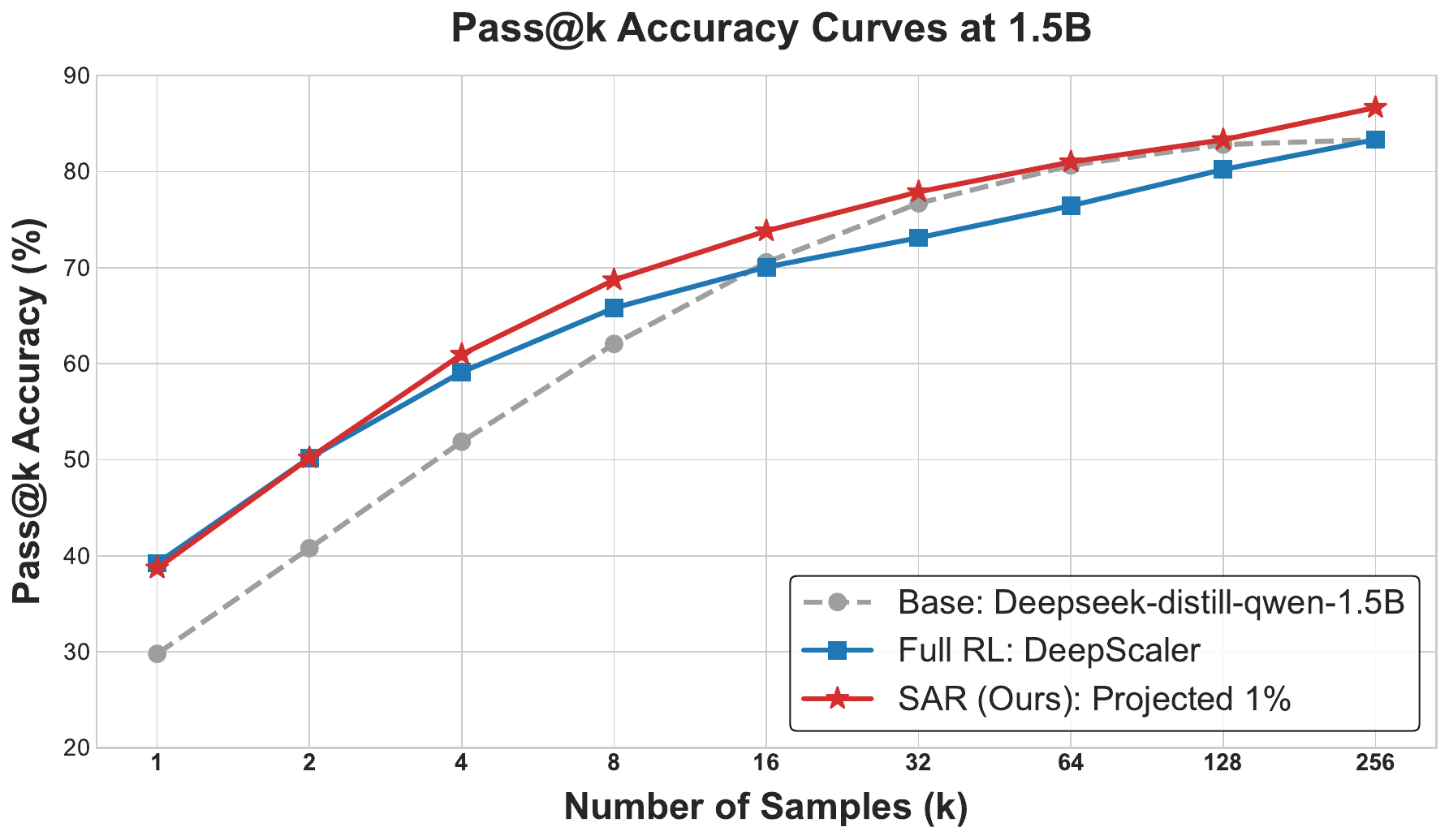}
         \caption{DeepScaleR-1.5B}
         \label{fig:passk_deepscaler}
     \end{subfigure}
     \hfill
     \begin{subfigure}[b]{0.48\textwidth}
         \centering
         \includegraphics[width=\linewidth]{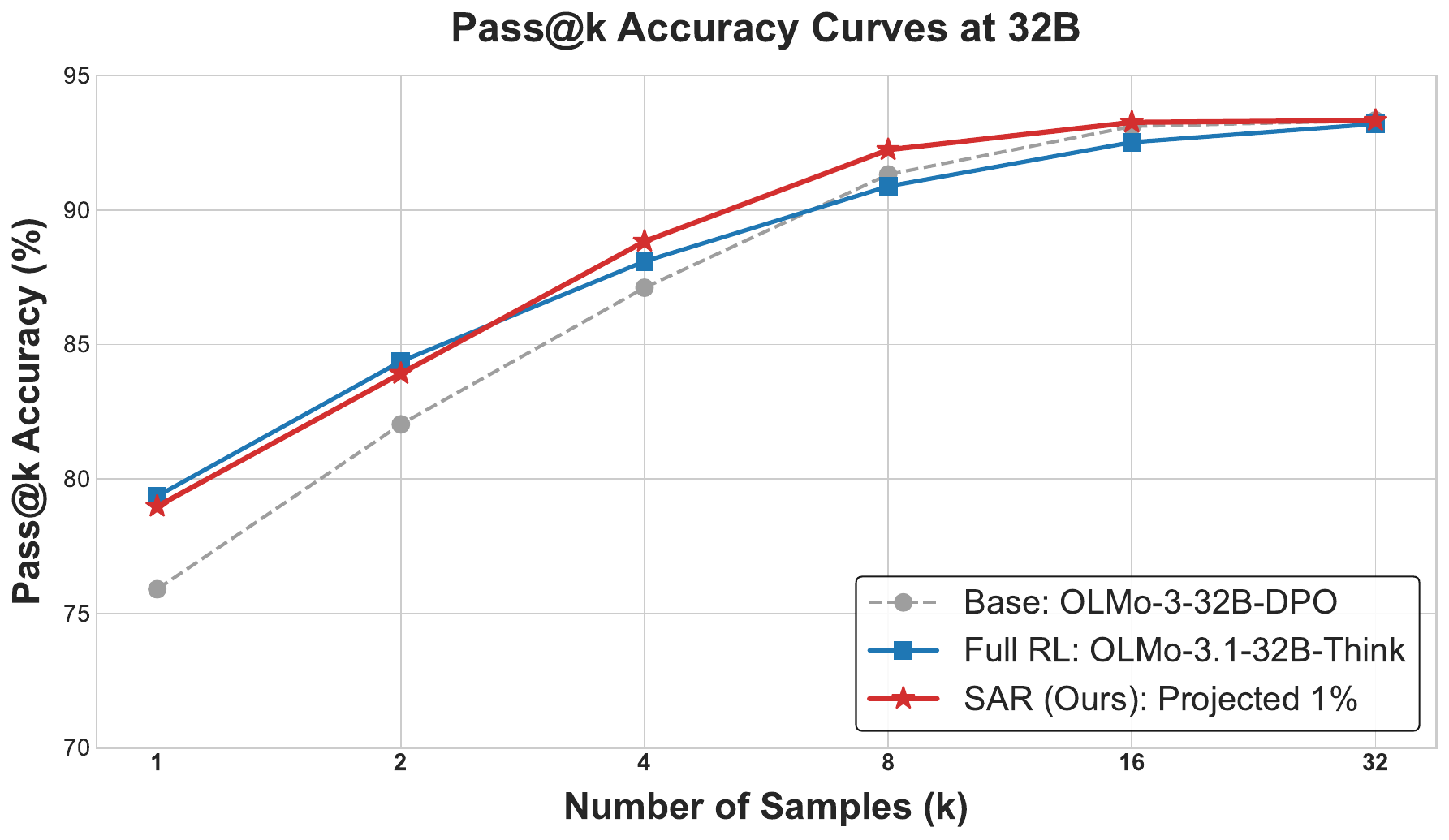}
         \caption{OLMo-3.1-32B-Think}
         \label{fig:passk_olmo}
     \end{subfigure}
     
     \caption{Pass@$k$ scaling curves on AIME 2024. Unconstrained RL baselines exhibit early saturation. In contrast, SAR models extract a compact reasoning-effective update, improving useful exploration and scaling momentum.}
     \label{fig:passk_scaling_combined}
\end{figure}

RL post-training often improves Pass@1 but can saturate under larger sampling budgets, limiting the benefit of test-time scaling. Under our spectral view, this suggests that the full RL update may contain both reasoning-effective directions and residual directions that narrow useful exploration. SAR tests this hypothesis by retaining the compact spectral rewiring component while removing the orthogonal residual update.

\paragraph{Experimental Setup.}
We evaluate this effect on two model families: DeepScaleR (1.5B) and OLMo-3.1-32B-Think. For each AIME 2024 problem, we generate 256 rollouts and compute Pass@$k$ for $k \in [1,128]$ using the same decoding configuration across all compared models \cite{chen2021evaluating,brown2024large}. To complement the Pass@$k$ curves, we also report a coverage-based metric following prior reasoning-RL analyses: a problem is counted as covered if at least one rollout among a fixed sampling budget produces the correct final answer. This metric measures useful exploration, since it counts only diversity that leads to a verified solution \cite{wang2022selfconsistency}.

\paragraph{Results.} 
Figure~\ref{fig:passk_scaling_combined} shows the Pass@$k$ scaling curves on AIME 2024. The full RL baselines exhibit early saturation: their performance improves at small $k$ but gains diminish at larger sampling budgets. In contrast, SAR-projected models continue to benefit from additional samples and overtake the corresponding full RL baselines at small-to-moderate $k$ values. For DeepScaleR and OLMo-3.1-32B-Think, SAR crosses the full RL curve at $k=2$ and $k=4$, respectively, and remains better at larger $k$.

Importantly, SAR preserves the average reasoning gains of RL while achieving \textbf{stronger large-$k$ scaling}. The same trend appears at the problem-coverage level: with 256 rollouts on AIME 2024, SAR covers 26/30 problems, compared with 25/30 for the full RL model (Appendix Table~\ref{tab:coverage_256}). Diagnostic controls further show that this gain is not explained by low-rank extraction alone, as no-projection, random, diagonal-only, and off-diagonal-only variants perform worse than SAR (Appendix~\ref{app:no_projection_control} and Appendix~\ref{app:method_ablations}).

\subsection{Additional In-House Agentic Coding Validation}
\label{subsec:inhouse_agentic_coding}

To evaluate SAR as an application-oriented editing method beyond public mathematical reasoning benchmarks, we further test it on the in-house model using open agentic coding benchmarks, as shown in Table~\ref{tab:inhouse_agentic_coding}. Starting from the same coding post-training delta, the top-1\% spectral-rank projection improves six of seven open agentic coding benchmarks, with a \textbf{+2.52\% improvement on average} and gains up to 25.10\%. This result shows that spectral rewiring can extract useful post-training signal and transfer to applied coding workflows.

\begin{table}[htbp]
\centering
\caption{Additional single-domain validation on the in-house model using open agentic coding benchmarks. SAR uses a top-1\% spectral-rank projection of the coding post-training delta.}
\label{tab:inhouse_agentic_coding}
\resizebox{\textwidth}{!}{%
\begin{tabular}{llcccc}
\toprule
\textbf{Benchmark} & \textbf{Agent} & \textbf{Baseline (In-house)} & \textbf{SAR (1\% Rank)} & \textbf{Relative Change} \\
\midrule
SWE-Bench-Pro\cite{swepro} & CodeAct\cite{openhands} & 0.297 & \textbf{0.327} & \textbf{+10.21\%} \\
SWE-Bench\cite{swe} & ClaudeCode & 0.580 & \textbf{0.590} & \textbf{+1.72\%} \\
SWE-Bench\cite{swe} & CodeAct\cite{openhands} & 0.566 & \textbf{0.574} & \textbf{+1.41\%} \\
MSWE-Bench\cite{mswe} & ClaudeCode & 0.255 & \textbf{0.313} & \textbf{+22.86\%} \\
MSWE-Bench\cite{mswe} & CodeAct\cite{openhands} & 0.271 & 0.270 & -0.33\% \\
TerminalBench 2.0\cite{terminal} & Terminus\cite{terminal} & 0.194 & \textbf{0.243} & \textbf{+25.10\%} \\
TerminalBench 1.0\cite{terminal} & Terminus\cite{terminal} & 0.323 & \textbf{0.344} & \textbf{+6.73\%} \\
\bottomrule
\end{tabular}%
}
\end{table}

\begin{takeawaybox}
\begin{itemize}
    \item SAR preserves full-RL-level math reasoning across model scales using a highly compact spectral update, showing that the reasoning-effective component of RL is strongly concentrated in the pretrained spectral geometry.
    \item The same compact update also improves high-$k$ reasoning behavior without retraining the model, changing the reward objective, or adding new rollouts.
    \item The validation on the in-house model shows that SAR can also improve applied coding performance beyond public math benchmarks.
    \item The main results and ablations show that the reasoning-effective parameters primarily rewire the pretrained spectral space, providing parameter-level geometric evidence that RL incentivizes capabilities already latent in the base model.
    \item These results suggest a training-free route to improve useful exploration after RL, and point toward future training recipes that optimize directly within the reasoning-effective spectral geometry.
\end{itemize}
\end{takeawaybox}

\section{Cross-Domain Generalization via Spectral Reasoning Rewiring}\label{sec:cross_domain}

Having shown that SAR preserves single-domain reasoning gains and improves large-$k$ exploration, we next study cross-domain generalization. Training or merging models across mathematics, coding, and instruction following can introduce domain-specific update components that interfere with each other. Under our geometric view, the fundamental reasoning component should lie in the base spectral manifold; SAR therefore projects reasoning updates onto this manifold to retain shared structure and filter incompatible residual directions. We evaluate this principle in two settings: jointly trained multi-domain RL models (\S \ref{subsec:mix_rl}) and post-hoc merging of independently trained domain experts (\S \ref{subsec:model_merging}).

\begin{figure}[htbp]
\centering\includegraphics[width=1.0\linewidth]{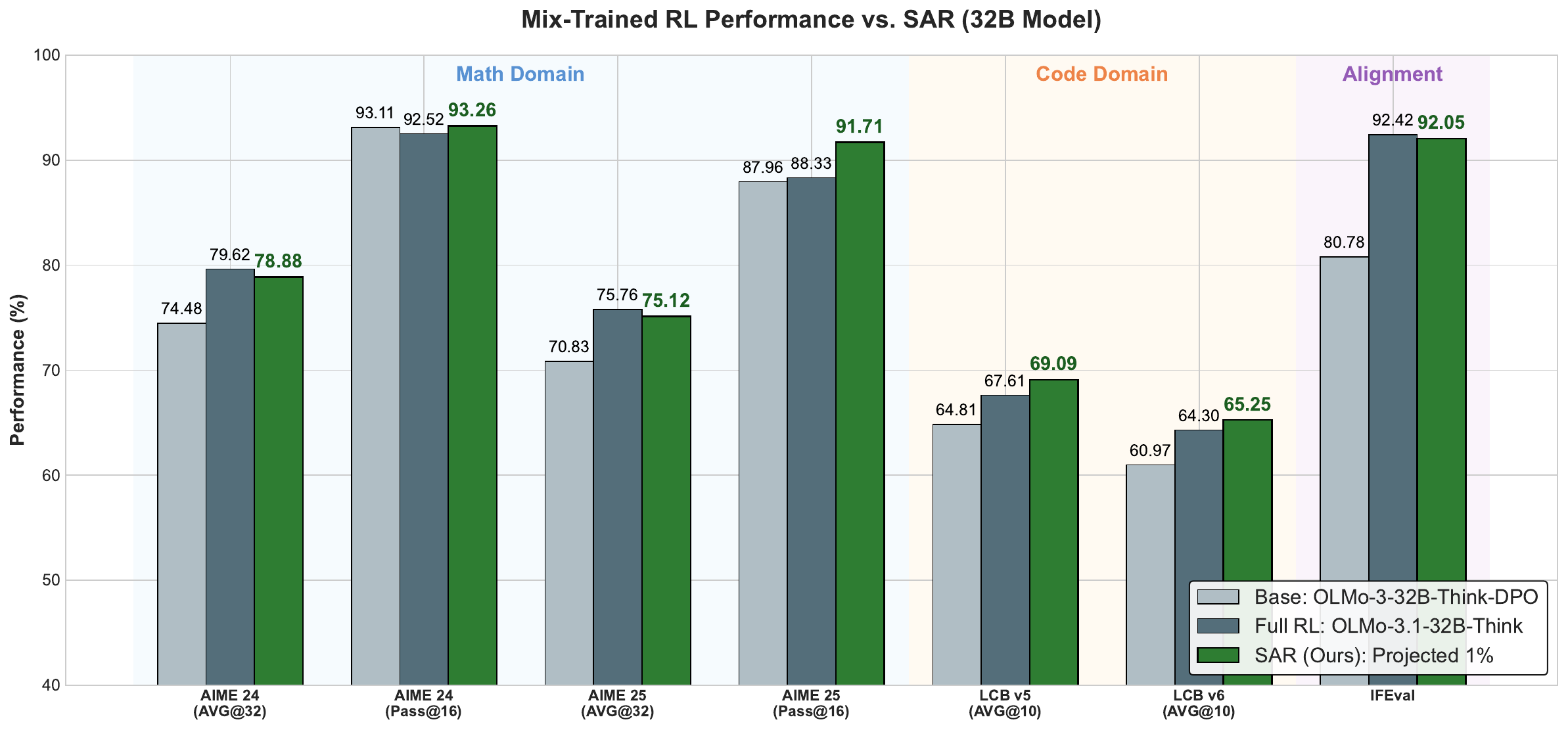}
\caption{Cross-domain performance evaluation on the OLMo-3-32B scale. By extracting reasoning-effective parameters, SAR reduces cross-domain interference, improves LiveCodeBench v5/v6, maintains instruction following (IFEval) and math reasoning (AIME 24), and enhances exploration capability.}
\label{fig:radar_mix_rl}
\end{figure}

\subsection{Purification: Reducing Domain Interference in Mix-RL}\label{subsec:mix_rl}
We first evaluate this projection principle in a jointly trained Mix-RL model, where multiple capabilities are optimized within the same full-parameter update.

\paragraph{Experimental Setup.} We evaluate the OLMo-3-32B series, which achieves state-of-the-art open-source performance. We use OLMo-3-32B-DPO as the base model and OLMo-3.1-32B-Think as the Mix-RL baseline, which is jointly trained on math, coding, instruction following, and chat. We apply SAR with the same top-1\% spectral-rank setting used in the single-domain experiments. Evaluation spans rigorous mathematical reasoning (AIME 24/25), algorithmic coding (LiveCodeBench v5/v6) \cite{jain2024livecodebench}, and general instruction following (IFEval) \cite{zhou2023instruction}.

\paragraph{Results.} As illustrated in Figure \ref{fig:radar_mix_rl}, extracting the reasoning-effective parameters via SAR improves performance on competitive coding while maintaining capacity in math and instruction following. 
\begin{itemize}
    \item \textit{Coding:} The projected model outperforms the full RL baseline on LCB v5 (from 67.61\% to 69.09\%) and v6 (from 64.30\% to 65.25\%). We observe consistent improvements in Pass@$k$ on LCB v5 and v6; Appendix~\ref{app:no_projection_control} provides an additional multi-domain control on LCB v5.
    \item \textit{Math:} The peak math reasoning performance is preserved with a marginal variance of less than 1\% in AVG@32, while exploration capability is significantly improved (e.g., from 88.33\% to 91.71\% on AIME 25 Pass@16).
    \item \textit{Instruction Following:} General instruction-following capability is also retained, with IFEval changed only slightly from 92.42\% to 92.05\%. 
\end{itemize}

With less than 1\% of the model parameters represented in the rewiring matrix $M$, SAR improves LiveCodeBench, an execution-sensitive coding benchmark, while preserving instruction following and mathematical reasoning. SAR also improves math and coding Pass@$k$ at larger sampling budgets, indicating that the projected model retains the accuracy gains of RL while recovering additional exploration capacity. These improvements are difficult to explain as lossy compression alone: reducing the update would typically be expected to preserve or degrade downstream performance, rather than improve a challenging held-out domain. We therefore interpret the LiveCodeBench gain as evidence that the full Mix-RL update contains domain-specific residual directions that interfere across tasks, while SAR exposes a compact spectral rewiring that is more compatible across domains.

\subsection{Merging: Improving Cross-Domain Expert Merging}\label{subsec:model_merging}

We next evaluate whether spectral rewiring can improve cross-domain model merging, focusing on math and code experts. Directly merging independently trained domain experts often degrades performance because their raw task deltas contain incompatible residual directions. SAR instead filters one expert through the pretrained spectral geometry before merging it with another.

\paragraph{Experimental Setup.} We consider a scenario involving two domain experts independently derived via RL from the same base model. At the 1.5B scale (Base: DeepSeek-Distill-Qwen-1.5B) \cite{qwen2024qwen25,yang2024qwen25math}, we use DeepScaleR (Math) and Archer-Code (Code) \cite{wang2025stabilizing}. At the 14B scale (Base: DeepSeek-Distill-Qwen-14B), we pair OpenReasoner (Math) \cite{hu2025openreasonerzero} with DeepCoder (Code) \cite{together2025deepcoder}. We use SAR to extract a compact mathematical reasoning rewiring from the math expert and merge these reasoning-effective parameters with the code expert. We benchmark against established weight interpolation methods including Task Arithmetic (TA), TIES, and DARE+TIES.

\paragraph{Results.}
Table~\ref{tab:model_merging} reports zero-shot math-code merging results. Task Arithmetic directly averages the full math and code RL updates,
and therefore serves as a simple averaging baseline. Its degradation, especially on the reasoning metric, suggests that naively merging full RL updates can introduce cross-domain interference.

At both 1.5B and 14B scales, SAR produces merged models that \textbf{exceed the best single-domain experts} on the primary math and code AVG metrics. At 1.5B, SAR improves AIME AVG@32 to 43.44\% and LCB AVG@8 to 32.25\%, surpassing the best math expert (40.31\%) and code expert (31.80\%) simultaneously. At 14B, SAR again surpasses the best math expert on AIME AVG@32 (74.38\% vs. 74.27\%) and improves beyond the code expert on LCB AVG@8 (63.40\% vs. 61.00\%), while matching the best AIME Pass@32. Although DARE+TIES obtains a slightly higher LCB score, SAR provides stronger balanced transfer in this setting by improving coding performance while preserving the math expert's advantage.

These results indicate that the reasoning rewiring extracted by SAR is more compatible for merging than the full RL update. In contrast to magnitude-based merging methods, SAR filters the update through the pretrained spectral geometry before merging, reducing interference with the code expert.

\begin{table*}[t]
\centering
\renewcommand{\arraystretch}{1.1}
\caption{Cross-Domain Generalization via Model Merging. By extracting and integrating compact reasoning-effective parameters, SAR outperforms naive interpolation and improves cross-domain transfer.}
\label{tab:model_merging}
\resizebox{0.8\textwidth}{!}{
\begin{tabular}{llccc}
\toprule
\multirow{2}{*}{\textbf{Scale}} & \multirow{2}{*}{\textbf{Model / Method}} & \multicolumn{2}{c}{\textbf{Math (AIME 24)}} & \textbf{Code (LCB)} \\
\cmidrule(lr){3-4} \cmidrule(lr){5-5}
& & \textbf{AVG@32} & \textbf{Pass@32} & \textbf{AVG@8} \\
\midrule

\multirow{9}{*}{\textbf{1.5B}} 
& \multicolumn{4}{l}{\textit{\color{gray}Single-Domain References}} \\
& Base: Distill-Qwen-1.5B & 31.67\% & 76.67\% & 23.00\% \\
& Math Expert: DeepScaleR & \textbf{40.31\%} & \textbf{76.67\%} & 27.20\% \\
& Code Expert: Archer-Code & 39.48\% & 76.67\% & \textbf{31.80\%} \\
\cmidrule(lr){2-5}
& \textit{Best Single-Domain Expert} & \textit{40.31\%} & \textit{76.67\%} & \textit{31.80\%} \\
\cmidrule(lr){2-5}
& \multicolumn{4}{l}{\textit{\color{gray}Cross-Domain Merging}} \\
& Task Arithmetic & 36.25\% & 63.33\% & \underline{31.80\%} \\
& TIES Merging & \underline{42.08\%} & \underline{73.33\%} & 31.40\% \\
& DARE + TIES & 40.05\% & \underline{73.33\%} & 31.70\% \\
& \cellcolor{gray!15}\textbf{Projected (Ours)} & \cellcolor{gray!15}\textbf{43.44\%} & \cellcolor{gray!15}\textbf{76.67\%} & \cellcolor{gray!15}\textbf{32.25\%} \\
\midrule

\multirow{9}{*}{\textbf{14B}} 
& \multicolumn{4}{l}{\textit{\color{gray}Single-Domain References}} \\
& Base: Distill-Qwen-14B & 67.71\% & 90.00\% & 54.20\% \\
& Math Expert: OR-Zero & \textbf{74.27\%} & \textbf{93.33\%} & 56.70\% \\
& Code Expert: DeepCoder & 71.04\% & 86.67\% & \textbf{61.00\%} \\
\cmidrule(lr){2-5}
& \textit{Best Single-Domain Expert} & \textit{74.27\%} & \textit{93.33\%} & \textit{61.00\%} \\
\cmidrule(lr){2-5}
& \multicolumn{4}{l}{\textit{\color{gray}Cross-Domain Merging}} \\
& Task Arithmetic & 71.04\% & \underline{93.33\%} & 54.60\% \\
& TIES Merging & \underline{73.54\%} & \underline{93.33\%} & 63.30\% \\
& DARE + TIES & 73.06\% & 90.00\% & \textbf{64.30\%} \\
& \cellcolor{gray!15}\textbf{Projected (Ours)} & \cellcolor{gray!15}\textbf{74.38\%} & \cellcolor{gray!15}\textbf{93.33\%} & \cellcolor{gray!15}\underline{63.40\%} \\
\bottomrule
\end{tabular}
}
\end{table*}

\begin{takeawaybox}
\begin{itemize}
    \item In Mix-RL, SAR releases suppressed cross-domain capability, improving coding and large-$k$ exploration while preserving math reasoning and instruction following.
    \item In expert merging, SAR produces merged models that surpass the best single-domain experts on the primary math and code AVG metrics at both 1.5B and 14B scales.
    \item These gains are consistent with our core view: the transferable reasoning component is concentrated in the pretrained spectral manifold, while residual directions outside this geometry can amplify domain interference.
    \item Spectral projection therefore makes reasoning updates more compatible across domains by retaining the shared rewiring component and filtering less compatible task-specific directions.
    \item This suggests a practical path for building stronger multi-domain reasoning models: post-training updates can be purified before deployment or merged across experts without directly combining all dense task deltas.
\end{itemize}
\end{takeawaybox}

\section{Discussion}

Across single-domain extraction, Mix-RL purification, and expert merging, SAR preserves or improves key downstream performance after projecting dense RL updates into the pretrained spectral coordinates. These results support our central insight: a large part of the reasoning-effective change induced by RL is not arbitrary parameter drift, but compact rewiring of capabilities already represented in the reference model's spectral geometry. This insight also defines the scope of the method. SAR is expected to be most effective when the reference model already contains the relevant reasoning ability or domain knowledge and RL primarily elicits or reorganizes it.

The main boundary of SAR is projection compatibility: how much of an RL update can be represented as a reorganization of knowledge already encoded in the reference model. Appendix~\ref{app:boundary_cases} analyzes three representative cases. In heavily optimized reasoning RL, projecting JustRL after more than 4000 training steps still recovers a strong AIME 2024 score of about 43\%, comparable to DeepScaleR-level reasoning, but does not match the final RL model. This suggests that the spectral component captures substantial elicitation, while the remaining gap may reflect task-specific rewriting outside the reference manifold. In code RL, direct RL from a base model is less projection-compatible because the update may learn code formatting, tool-interface behavior, and execution-oriented conventions, rather than only recombining existing knowledge and skills. After code SFT establishes this domain knowledge, the subsequent RL update becomes more compatible with spectral projection. PPO-style training with an external critic creates a related boundary: distribution shift between the critic and actor can induce preferences that are not well aligned with the reference model's spectral manifold. Thus, projection compatibility provides a diagnostic for whether RL primarily reorganizes latent capabilities within the reference manifold or acquires behavior beyond it.
\section{Conclusion}

We presented a first-principles geometric view of RL post-training in large language models. By analyzing updates in the pretrained singular-vector coordinates, we show that much of the reasoning-effective change is captured by compact spectral rewiring, consistent with the view that RL reorganizes latent capabilities already represented in the base model. Subspace-Aligned Rewiring (SAR) turns this insight into a post-hoc editing method that preserves full-RL-level reasoning, improves high-$k$ exploration, reduces multi-domain interference, and strengthens expert model merging. These results position spectral rewiring as both a lens for understanding post-training updates and a practical tool for improving their downstream utility.

    
    \clearpage
    \bibliographystyle{unsrtnat}
    \bibliography{refs}

\begin{thebibliography}{48}
\providecommand{\natexlab}[1]{#1}
\providecommand{\url}[1]{\texttt{#1}}
\expandafter\ifx\csname urlstyle\endcsname\relax
  \providecommand{\doi}[1]{doi: #1}\else
  \providecommand{\doi}{doi: \begingroup \urlstyle{rm}\Url}\fi

\bibitem[Shao et~al.(2024)Shao, Wang, Zhu, Xu, Song, Bi, Zhang, Zhang, Li, Wu, et~al.]{shao2024deepseekmath}
Zhihong Shao, Peiyi Wang, Qihao Zhu, Runxin Xu, Junxiao Song, Xiao Bi, Haowei Zhang, Mingchuan Zhang, YK~Li, Y~Wu, et~al.
\newblock Deepseekmath: Pushing the limits of mathematical reasoning in open language models.
\newblock \emph{arXiv preprint arXiv:2402.03300}, 2024.

\bibitem[Guo et~al.(2025)Guo, Yang, Zhang, Song, Zhang, Xu, Zhu, Ma, Wang, Bi, et~al.]{guo2025deepseekr1}
Daya Guo, Dejian Yang, Haowei Zhang, Junxiao Song, Ruoyu Zhang, Runxin Xu, Qihao Zhu, Shirong Ma, Peiyi Wang, Xiao Bi, et~al.
\newblock Deepseek-r1: Incentivizing reasoning capability in llms via reinforcement learning.
\newblock \emph{arXiv preprint arXiv:2501.12948}, 2025.

\bibitem[Cobbe et~al.(2021)Cobbe, Kosaraju, Bavarian, Chen, Jun, Kaiser, Plappert, Tworek, Hilton, Nakano, et~al.]{cobbe2021training}
Karl Cobbe, Vineet Kosaraju, Mohammad Bavarian, Mark Chen, Heewoo Jun, Lukasz Kaiser, Matthias Plappert, Jerry Tworek, Jacob Hilton, Reiichiro Nakano, et~al.
\newblock Training verifiers to solve math word problems.
\newblock \emph{arXiv preprint arXiv:2110.14168}, 2021.

\bibitem[Lightman et~al.(2023)Lightman, Kosaraju, Burda, Edwards, Baker, Lee, Leike, Schulman, Sutskever, and Cobbe]{lightman2023lets}
Hunter Lightman, Vineet Kosaraju, Yura Burda, Harrison Edwards, Bowen Baker, Teddy Lee, Jan Leike, John Schulman, Ilya Sutskever, and Karl Cobbe.
\newblock Let's verify step by step.
\newblock \emph{arXiv preprint arXiv:2305.20050}, 2023.

\bibitem[Kirk et~al.(2023)Kirk, Vidgen, Rottger, and Hale]{kirk2023understanding}
Hannah~Rose Kirk, Bertie Vidgen, Paul Rottger, and Scott~A. Hale.
\newblock Understanding the effects of rlhf on llm generalisation and diversity.
\newblock \emph{arXiv preprint arXiv:2310.06452}, 2023.

\bibitem[Moskovitz et~al.(2023)Moskovitz, Singh, Strouse, Sandholm, Salakhutdinov, Dragan, and McAleer]{moskovitz2023confronting}
Ted Moskovitz, Archit Singh, DJ~Strouse, Tuomas Sandholm, Ruslan Salakhutdinov, Anca Dragan, and Stephen McAleer.
\newblock Confronting reward model overoptimization with constrained rlhf.
\newblock \emph{arXiv preprint arXiv:2310.04373}, 2023.

\bibitem[Brown et~al.(2024)Brown, Juravsky, Ehrlich, Clark, Le, Re, and Mirhoseini]{brown2024large}
Bradley C.~A. Brown, Jordan Juravsky, Ryan Ehrlich, Ronald Clark, Quoc~V. Le, Christopher Re, and Azalia Mirhoseini.
\newblock Large language monkeys: Scaling inference compute with repeated sampling.
\newblock \emph{arXiv preprint arXiv:2407.21787}, 2024.

\bibitem[Snell et~al.(2024)Snell, Lee, Xu, and Kumar]{snell2024scaling}
Charlie Snell, Jaehoon Lee, Kelvin Xu, and Aviral Kumar.
\newblock Scaling llm test-time compute optimally can be more effective than scaling model parameters.
\newblock \emph{arXiv preprint arXiv:2408.03314}, 2024.

\bibitem[Ilharco et~al.(2022)Ilharco, Ribeiro, Wortsman, Gururangan, Schmidt, Hajishirzi, and Farhadi]{ilharco2022editing}
Gabriel Ilharco, Marco~Tulio Ribeiro, Mitchell Wortsman, Suchin Gururangan, Ludwig Schmidt, Hannaneh Hajishirzi, and Ali Farhadi.
\newblock Editing models with task arithmetic.
\newblock \emph{arXiv preprint arXiv:2212.04089}, 2022.

\bibitem[Yadav et~al.(2023)Yadav, Tam, Choshen, Raffel, and Bansal]{yadav2023ties}
Prateek Yadav, Derek Tam, Leshem Choshen, Colin Raffel, and Mohit Bansal.
\newblock Ties-merging: Resolving interference when merging models.
\newblock \emph{arXiv preprint arXiv:2306.01708}, 2023.

\bibitem[Yu et~al.(2023)Yu, Yu, Yu, Huang, and Li]{yu2023language}
Le~Yu, Bowen Yu, Haiyang Yu, Fei Huang, and Yongbin Li.
\newblock Language models are super mario: Absorbing abilities from homologous models as a free lunch.
\newblock \emph{arXiv preprint arXiv:2311.03099}, 2023.

\bibitem[Han et~al.(2015)Han, Mao, and Dally]{han2015deep}
Song Han, Huizi Mao, and William~J. Dally.
\newblock Deep compression: Compressing deep neural networks with pruning, trained quantization and huffman coding.
\newblock \emph{arXiv preprint arXiv:1510.00149}, 2015.

\bibitem[Yuan et~al.(2023)Yuan, Shang, Song, Wu, Yan, and Sun]{yuan2023asvd}
Zhihang Yuan, Yuzhang Shang, Yue Song, Qiang Wu, Yan Yan, and Guangyu Sun.
\newblock Asvd: Activation-aware singular value decomposition for compressing large language models.
\newblock \emph{arXiv preprint arXiv:2312.05821}, 2023.

\bibitem[Wang et~al.(2024)Wang, Zheng, Wan, and Zhang]{wang2024svdllm}
Xin Wang, Yu~Zheng, Zhongwei Wan, and Mi~Zhang.
\newblock Svd-llm: Truncation-aware singular value decomposition for large language model compression.
\newblock \emph{arXiv preprint arXiv:2403.07378}, 2024.

\bibitem[Halford et~al.(1998)Halford, Wilson, and Phillips]{halford1998processing}
Graeme~S. Halford, William~H. Wilson, and Steven Phillips.
\newblock Processing capacity defined by relational complexity: Implications for comparative, developmental, and cognitive psychology.
\newblock \emph{Behavioral and Brain Sciences}, 21\penalty0 (6):\penalty0 803--831, 1998.

\bibitem[Santoro et~al.(2017)Santoro, Raposo, Barrett, Malinowski, Pascanu, Battaglia, and Lillicrap]{santoro2017simple}
Adam Santoro, David Raposo, David G.~T. Barrett, Mateusz Malinowski, Razvan Pascanu, Peter Battaglia, and Timothy Lillicrap.
\newblock A simple neural network module for relational reasoning.
\newblock \emph{arXiv preprint arXiv:1706.01427}, 2017.

\bibitem[Battaglia et~al.(2018)Battaglia, Hamrick, Bapst, Sanchez-Gonzalez, Zambaldi, Malinowski, Tacchetti, Raposo, Santoro, Faulkner, et~al.]{battaglia2018relational}
Peter~W. Battaglia, Jessica~B. Hamrick, Victor Bapst, Alvaro Sanchez-Gonzalez, Vinicius Zambaldi, Mateusz Malinowski, Andrea Tacchetti, David Raposo, Adam Santoro, Ryan Faulkner, et~al.
\newblock Relational inductive biases, deep learning, and graph networks.
\newblock \emph{arXiv preprint arXiv:1806.01261}, 2018.

\bibitem[Luo et~al.(2025)Luo, Tan, Wong, Shi, Tang, Roongta, Cai, Luo, Li, Yang, et~al.]{luo2025deepscaler}
Michael Luo, Sijun Tan, Justin Wong, Xiaoxiang Shi, William~Y. Tang, Manan Roongta, Colin Cai, Jeffrey Luo, Tianjun Li, Li~Erran Yang, et~al.
\newblock Deepscaler: Surpassing o1-preview with a 1.5b model by scaling rl.
\newblock \emph{arXiv preprint arXiv:2502.01682}, 2025.

\bibitem[{POLARIS Project}(2025)]{polaris2025preview}
{POLARIS Project}.
\newblock {POLARIS}: A post-training recipe for scaling reinforcement learning on advanced reasoning models.
\newblock \url{https://hkunlp.github.io/blog/2025/Polaris/}, 2025.
\newblock Accessed 2026-05-15.

\bibitem[{AI2}(2025)]{allenai2025olmo3}
{AI2}.
\newblock {OLMo 3}: A family of open language models.
\newblock \emph{arXiv preprint arXiv:2512.13961}, 2025.

\bibitem[Hu et~al.(2025)Hu, Wu, Fu, Chen, Xu, Zhu, Pei, Zhong, Zheng, Chen, et~al.]{hu2025openreasonerzero}
Jian Hu, Xibin Wu, Weixun Fu, Xinyu Chen, Chen Xu, Weizhi Zhu, Jiaxin Pei, Zixuan Zhong, Jiawei Zheng, Zheng Chen, et~al.
\newblock Open-reasoner-zero: An open source approach to scaling reinforcement learning on the base model.
\newblock \emph{arXiv preprint arXiv:2503.24290}, 2025.

\bibitem[Chen et~al.(2021)Chen, Tworek, Jun, Yuan, Pinto, Kaplan, Edwards, Burda, Joseph, Brockman, et~al.]{chen2021evaluating}
Mark Chen, Jerry Tworek, Heewoo Jun, Qiming Yuan, Henrique Ponde de~Oliveira Pinto, Jared Kaplan, Harri Edwards, Yuri Burda, Nicholas Joseph, Greg Brockman, et~al.
\newblock Evaluating large language models trained on code.
\newblock \emph{arXiv preprint arXiv:2107.03374}, 2021.

\bibitem[Wang et~al.(2022)Wang, Wei, Schuurmans, Le, Chi, Narang, Chowdhery, and Zhou]{wang2022selfconsistency}
Xuezhi Wang, Jason Wei, Dale Schuurmans, Quoc Le, Ed~Chi, Sharan Narang, Aakanksha Chowdhery, and Denny Zhou.
\newblock Self-consistency improves chain of thought reasoning in language models.
\newblock \emph{arXiv preprint arXiv:2203.11171}, 2022.

\bibitem[Deng et~al.(2025)Deng, Da, Pan, He, Ide, Garg, Lauffer, Park, Pasari, Rane, Sampath, Krishnan, Kundurthy, Hendryx, Wang, Zhang, Jacobson, Liu, and Kenstler]{swepro}
Xiang Deng, Jeff Da, Edwin Pan, Yan He, Charles Ide, Kanak Garg, Niklas Lauffer, Andrew Park, Nitin Pasari, Chetan Rane, Karmini Sampath, Maya Krishnan, Srivatsa Kundurthy, Sean~M. Hendryx, Zifan Wang, Chen Bo~Calvin Zhang, Noah Jacobson, Bing Liu, and Brad Kenstler.
\newblock Swe-bench pro: Can ai agents solve long-horizon software engineering tasks?
\newblock 2025.
\newblock URL \url{https://api.semanticscholar.org/CorpusID:281421060}.

\bibitem[Wang et~al.(2025{\natexlab{a}})Wang, Li, Song, Xu, Tang, Zhuge, Pan, Song, Li, Singh, Tran, Li, Ma, Zheng, Qian, Shao, Muennighoff, Zhang, Hui, Lin, Brennan, Peng, Ji, and Neubig]{openhands}
Xingyao Wang, Boxuan Li, Yufan Song, Frank~F Xu, Xiangru Tang, Mingchen Zhuge, Jiayi Pan, Yueqi Song, Bowen Li, Jaskirat Singh, Hoang Tran, Fuqiang Li, Ren Ma, Mingzhang Zheng, Bill Qian, Daniel Shao, Niklas Muennighoff, Yizhe Zhang, Binyuan Hui, Junyang Lin, Robert Brennan, Hao Peng, Heng Ji, and Graham Neubig.
\newblock Openhands: An open platform for ai software developers as generalist agents.
\newblock In Y.~Yue, A.~Garg, N.~Peng, F.~Sha, and R.~Yu, editors, \emph{International Conference on Learning Representations}, volume 2025, pages 65882--65919, 2025{\natexlab{a}}.
\newblock URL \url{https://proceedings.iclr.cc/paper_files/paper/2025/file/a4b6ad6b48850c0c331d1259fc66a69c-Paper-Conference.pdf}.

\bibitem[Jimenez et~al.(2024)Jimenez, Yang, Wettig, Yao, Pei, Press, and Narasimhan]{swe}
Carlos~E Jimenez, John Yang, Alexander Wettig, Shunyu Yao, Kexin Pei, Ofir Press, and Karthik~R Narasimhan.
\newblock Swe-bench: Can language models resolve real-world github issues?
\newblock In \emph{The Twelfth International Conference on Learning Representations}, 2024.
\newblock URL \url{https://openreview.net/forum?id=VTF8yNQM66}.

\bibitem[Zan et~al.(2025)Zan, Huang, Liu, Chen, Zhang, Xin, Chen, Liu, Zhong, Li, Liu, Xiao, Chen, Zhang, Su, Liu, Long, Shen, and Xiang]{mswe}
Daoguang Zan, Zhirong Huang, Wei Liu, Hanwu Chen, Linhao Zhang, Shulin Xin, Lu~Chen, Qi~Liu, Xiaojian Zhong, Aoyan Li, Siyao Liu, Yongsheng Xiao, Liangqiang Chen, Yuyu Zhang, Jing Su, Tianyu Liu, Rui Long, Kai Shen, and Liang Xiang.
\newblock Multi-swe-bench: A multilingual benchmark for issue resolving, 2025.
\newblock URL \url{https://arxiv.org/abs/2504.02605}.

\bibitem[Merrill et~al.(2026)Merrill, Shaw, Carlini, Li, Raj, Bercovich, Shi, Shin, Walshe, Buchanan, Shen, Ye, Lin, Poulos, Wang, Nezhurina, Jitsev, Lu, Mastromichalakis, Xu, Chen, Liu, Zhang, Chen, Kashyap, Uslu, Li, Wu, Yan, Bian, Sharma, Sun, Dillmann, Anand, Lanpouthakoun, Koopah, Hu, Guha, Dreiman, Zhu, Krauth, Zhong, Muennighoff, Amanfu, Tan, Pimpalgaonkar, Aggarwal, Lin, Lan, Zhao, Liang, Wang, Wang, Zhou, Heineman, Liu, Trivedi, Yang, Lin, Shetty, Yang, Omi, Raoof, Li, Zhuo, Lin, Dai, Wang, Chai, Zhou, Wahdany, She, Hu, Dong, Zhu, Cui, Saiyed, Kolbeinsson, Hu, Rytting, Marten, Wang, Dimakis, Konwinski, and Schmidt]{terminal}
Mike~A. Merrill, Alexander~G. Shaw, Nicholas Carlini, Boxuan Li, Harsh Raj, Ivan Bercovich, Lin Shi, Jeong~Yeon Shin, Thomas Walshe, E.~Kelly Buchanan, Junhong Shen, Guanghao Ye, Haowei Lin, Jason Poulos, Maoyu Wang, Marianna Nezhurina, Jenia Jitsev, Di~Lu, Orfeas~Menis Mastromichalakis, Zhiwei Xu, Zizhao Chen, Yue Liu, Robert Zhang, Leon~Liangyu Chen, Anurag Kashyap, Jan-Lucas Uslu, Jeffrey Li, Jianbo Wu, Minghao Yan, Song Bian, Vedang Sharma, Ke~Sun, Steven Dillmann, Akshay Anand, Andrew Lanpouthakoun, Bardia Koopah, Changran Hu, Etash Guha, Gabriel H.~S. Dreiman, Jiacheng Zhu, Karl Krauth, Li~Zhong, Niklas Muennighoff, Robert Amanfu, Shangyin Tan, Shreyas Pimpalgaonkar, Tushar Aggarwal, Xiangning Lin, Xin Lan, Xuandong Zhao, Yiqing Liang, Yuanli Wang, Zilong Wang, Changzhi Zhou, David Heineman, Hange Liu, Harsh Trivedi, John Yang, Junhong Lin, Manish Shetty, Michael Yang, Nabil Omi, Negin Raoof, Shanda Li, Terry~Yue Zhuo, Wuwei Lin, Yiwei Dai, Yuxin Wang, Wenhao Chai, Shang Zhou, Dariush Wahdany, Ziyu She, Jiaming Hu, Zhikang Dong, Yuxuan Zhu, Sasha Cui, Ahson Saiyed, Arinbjörn Kolbeinsson, Jesse Hu, Christopher~Michael Rytting, Ryan Marten, Yixin Wang, Alex Dimakis, Andy Konwinski, and Ludwig Schmidt.
\newblock Terminal-bench: Benchmarking agents on hard, realistic tasks in command line interfaces, 2026.
\newblock URL \url{https://arxiv.org/abs/2601.11868}.

\bibitem[Jain et~al.(2024)Jain, Han, Gu, Li, Yan, Zhang, Wang, Solar-Lezama, Sen, and Stoica]{jain2024livecodebench}
Naman Jain, King Han, Alex Gu, Wen-Ding Li, Fanjia Yan, Tianjun Zhang, Sida Wang, Armando Solar-Lezama, Koushik Sen, and Ion Stoica.
\newblock Livecodebench: Holistic and contamination free evaluation of large language models for code.
\newblock \emph{arXiv preprint arXiv:2403.07974}, 2024.

\bibitem[Zhou et~al.(2023)Zhou, Lu, Mishra, Brahma, Basu, Luan, Zhou, and Hou]{zhou2023instruction}
Jeffrey Zhou, Tianjian Lu, Swaroop Mishra, Siddhartha Brahma, Sujoy Basu, Yi~Luan, Denny Zhou, and Le~Hou.
\newblock Instruction-following evaluation for large language models.
\newblock \emph{arXiv preprint arXiv:2311.07911}, 2023.

\bibitem[{Qwen Team}(2024)]{qwen2024qwen25}
{Qwen Team}.
\newblock Qwen2.5 technical report.
\newblock \emph{arXiv preprint arXiv:2412.15115}, 2024.

\bibitem[Yang et~al.(2024)Yang, Zhang, Hui, Gao, Yu, Li, Liu, Tu, Zhou, Lin, et~al.]{yang2024qwen25math}
An~Yang, Beichen Zhang, Binyuan Hui, Bofei Gao, Bowei Yu, Chengpeng Li, Dayiheng Liu, Jianhong Tu, Jingren Zhou, Junyang Lin, et~al.
\newblock Qwen2.5-math technical report: Toward mathematical expert model via self-improvement.
\newblock \emph{arXiv preprint arXiv:2409.12122}, 2024.

\bibitem[Wang et~al.(2025{\natexlab{b}})Wang, Liu, Zhang, Li, and Zhou]{wang2025stabilizing}
Jiakang Wang, Runze Liu, Fuzheng Zhang, Xiu Li, and Guorui Zhou.
\newblock Stabilizing knowledge, promoting reasoning: Dual-token constraints for rlvr, 2025{\natexlab{b}}.
\newblock URL \url{https://arxiv.org/abs/2507.15778}.

\bibitem[{Together AI}(2025)]{together2025deepcoder}
{Together AI}.
\newblock Deepcoder: A fully open-source 14b coder at o3-mini level.
\newblock \url{https://www.together.ai/blog/deepcoder}, 2025.
\newblock Accessed 2026-05-15.

\bibitem[Yue et~al.(2025)Yue, Chen, and Chen]{yue2025does}
Xiang Yue, Zhuo Chen, and Wenhu Chen.
\newblock Does reinforcement learning really incentivize reasoning capacity in llms beyond the base model?
\newblock \emph{arXiv preprint arXiv:2504.13837}, 2025.

\bibitem[Walder and Karkhanis(2025)]{walder2025pkpo}
Christian Walder and Deep Karkhanis.
\newblock Pass@k policy optimization: Solving harder reinforcement learning problems.
\newblock \emph{arXiv preprint arXiv:2505.15201}, 2025.

\bibitem[Chen et~al.(2025)Chen, Qin, Wu, Ling, Ye, Zhao, and Shi]{chen2025passktraining}
Zhipeng Chen, Xiaobo Qin, Youbin Wu, Yue Ling, Qinghao Ye, Wayne~Xin Zhao, and Guang Shi.
\newblock Pass@k training for adaptively balancing exploration and exploitation of large reasoning models.
\newblock \emph{arXiv preprint arXiv:2508.10751}, 2025.

\bibitem[Tajwar et~al.(2026)Tajwar, Zeng, Zhou, Song, Arora, Jiang, Schneider, Salakhutdinov, Feng, and Zanette]{tajwar2026maxrl}
Fahim Tajwar, Guanning Zeng, Yueer Zhou, Yuda Song, Daman Arora, Yiding Jiang, Jeff Schneider, Ruslan Salakhutdinov, Haiwen Feng, and Andrea Zanette.
\newblock Maximum likelihood reinforcement learning.
\newblock \emph{arXiv preprint arXiv:2602.02710}, 2026.

\bibitem[Peng et~al.(2025)Peng, Ren, Yu, Liu, and Wen]{peng2025simko}
Ruotian Peng, Yi~Ren, Zhouliang Yu, Weiyang Liu, and Yandong Wen.
\newblock Simko: Simple pass@k policy optimization.
\newblock \emph{arXiv preprint arXiv:2510.14807}, 2025.

\bibitem[Hu et~al.(2021)Hu, Shen, Wallis, Allen-Zhu, Li, Wang, Wang, and Chen]{hu2021lora}
Edward~J. Hu, Yelong Shen, Phillip Wallis, Zeyuan Allen-Zhu, Yuanzhi Li, Shean Wang, Lu~Wang, and Weizhu Chen.
\newblock Lora: Low-rank adaptation of large language models.
\newblock \emph{arXiv preprint arXiv:2106.09685}, 2021.

\bibitem[Liu et~al.(2024)Liu, Wang, Yin, Molchanov, Wang, Cheng, and Chen]{liu2024dora}
Shih-Yang Liu, Chien-Yi Wang, Hongxu Yin, Pavlo Molchanov, Yu-Chiang~Frank Wang, Kwang-Ting Cheng, and Min-Hung Chen.
\newblock Dora: Weight-decomposed low-rank adaptation.
\newblock \emph{arXiv preprint arXiv:2402.09353}, 2024.

\bibitem[Meng et~al.(2024)Meng, Wang, and Zhang]{meng2024pissa}
Fanxu Meng, Zhaohui Wang, and Muhan Zhang.
\newblock Pissa: Principal singular values and singular vectors adaptation of large language models.
\newblock \emph{arXiv preprint arXiv:2404.02948}, 2024.

\bibitem[Balazy et~al.(2024)Balazy, Banaei, Aberer, and Tabor]{balazy2024loraxs}
Klaudia Balazy, Mohammadreza Banaei, Karl Aberer, and Jacek Tabor.
\newblock Lora-xs: Low-rank adaptation with extremely small number of parameters.
\newblock \emph{arXiv preprint arXiv:2405.17604}, 2024.

\bibitem[Cai et~al.(2025)Cai, Cao, Xu, Yao, Huang, Tan, Zhang, Liu, and Fang]{cai2025predictability}
Yuchen Cai, Ding Cao, Xin Xu, Zijun Yao, Yuqing Huang, Zhenyu Tan, Benyi Zhang, Guiquan Liu, and Junfeng Fang.
\newblock On predictability of reinforcement learning dynamics for large language models.
\newblock \emph{arXiv preprint arXiv:2510.00553}, 2025.

\bibitem[Morris et~al.(2026)Morris, Mireshghallah, Ibrahim, and Mahloujifar]{morris2026learning}
John~X. Morris, Niloofar Mireshghallah, Mark Ibrahim, and Saeed Mahloujifar.
\newblock Learning to reason in 13 parameters.
\newblock \emph{arXiv preprint arXiv:2602.04118}, 2026.

\bibitem[Ren et~al.(2026)Ren, Lu, Wu, Zhao, Dai, Wu, Wang, and Zong]{ren2026enough}
Zixuan Ren, Jinliang Lu, Junhong Wu, Yang Zhao, Dai Dai, Hua Wu, Haifeng Wang, and Chengqing Zong.
\newblock Enough is as good as a feast: A comprehensive analysis of how reinforcement learning mitigates task conflicts in llms.
\newblock In \emph{International Conference on Learning Representations}, 2026.
\newblock URL \url{https://openreview.net/forum?id=N4l4Jp50R4}.

\bibitem[{JustRL Team}(2026)]{justrl2026scaling}
{JustRL Team}.
\newblock {JustRL}: Scaling a 1.5b llm with a simple rl recipe.
\newblock \url{https://iclr-blogposts.github.io/2026/blog/2026/justrl/}, 2026.
\newblock ICLR Blogposts 2026.

\bibitem[Wu et~al.(2026)Wu, Li, Zhang, Guo, Luo, Liu, Huang, Chu, Li, and Yang]{wu2026xcoder}
Jie Wu, Haoling Li, Xin Zhang, Jiani Guo, Jane Luo, Steven Liu, Yangyu Huang, Ruihang Chu, Scarlett Li, and Yujiu Yang.
\newblock {X-Coder}: Advancing competitive programming with fully synthetic tasks, solutions, and tests.
\newblock \emph{arXiv preprint arXiv:2601.06953}, 2026.

\end{thebibliography}

    \clearpage
    \beginappendix
    \section{Related Work}

\paragraph{Exploration and exploitation in outcome-reward RL.}
Recent work studies reasoning RL not only as a way to improve Pass@1, but also as a process that reshapes the exploration--exploitation trade-off of reasoning models \cite{yue2025does}. A related line of methods redesigns the RL objective to better optimize Pass@$k$-style behavior or test-time scaling efficiency \cite{walder2025pkpo,chen2025passktraining,tajwar2026maxrl,peng2025simko}. These approaches treat exploration primarily as a training-time objective design problem. SAR takes a \textbf{different view}: a completed RL update can already contain a compact high-$k$-effective component, but this component is entangled with residual off-manifold drift. By separating them in pretrained spectral coordinates, SAR improves Pass@$k$ through post-hoc model editing, without changing the RL objective, collecting new rollouts, or retraining the model.

\paragraph{LoRA and low-dimensional RL updates.}
Parameter-efficient adaptation and recent analyses of RL dynamics both suggest that LLM adaptation can be highly low-dimensional \cite{hu2021lora,liu2024dora,meng2024pissa,balazy2024loraxs,cai2025predictability,morris2026learning}. These works use low dimensionality as a training parameterization, compression principle, or predictability property. SAR instead identifies a more specific mechanism: the reasoning-effective component of a full RL update is not merely low-rank, but aligned with the pretrained model's spectral coordinates and structured as a compact rewiring matrix among pretrained components. This distinction enables post-hoc editing of a completed full-RL model and can surpass the RL baseline in high-$k$ exploration and multi-domain transfer, including at the 32B scale.

\paragraph{Model merging and interference.}
Model merging methods combine separately trained models by operating on weight-space task deltas, typically using arithmetic, magnitude, sign, or sparsity-based rules to reduce interference \cite{ilharco2022editing,yadav2023ties,yu2023language}. Recent analysis further shows that RL-trained models are more mergeable than SFT-trained models, but merged models can still degrade relative to their corresponding experts \cite{ren2026enough}. These works establish task conflict as a central obstacle in model composition. SAR addresses this conflict at a different level: before composition, it filters each RL update through the pretrained spectral manifold, extracting reasoning-effective rewiring while removing off-manifold residual drift. This enables positive-sum cross-domain composition where the merged model can surpass the best single-domain experts.

\section{Illustrative Example of Spectral Rewiring}
\label{app:mechanistic_example}

\textit{Triangle area calculation.}
Consider a geometric reasoning task that requires computing the area of a triangle from two side lengths and the included angle. In the pretrained base model, the relevant capabilities may be represented in separated spectral components: a right singular direction $v_1$ detects side-length information, another direction $v_2$ detects angle information, and an output direction $u_3$ corresponds to applying the sine-area formula. Under the original diagonal mapping $\Sigma$, neither $v_1$ nor $v_2$ alone is sufficient to activate $u_3$. SAR isolates the RL-induced off-diagonal entries $M_{3,1}$ and $M_{3,2}$, which route both premise directions into the same output direction. Thus, when the model detects both required premises, Equation~\ref{eq:rewired_forward} combines their signals to activate the appropriate multi-condition deduction.

\begin{figure}
    \centering
    \includegraphics[width=0.9\linewidth]{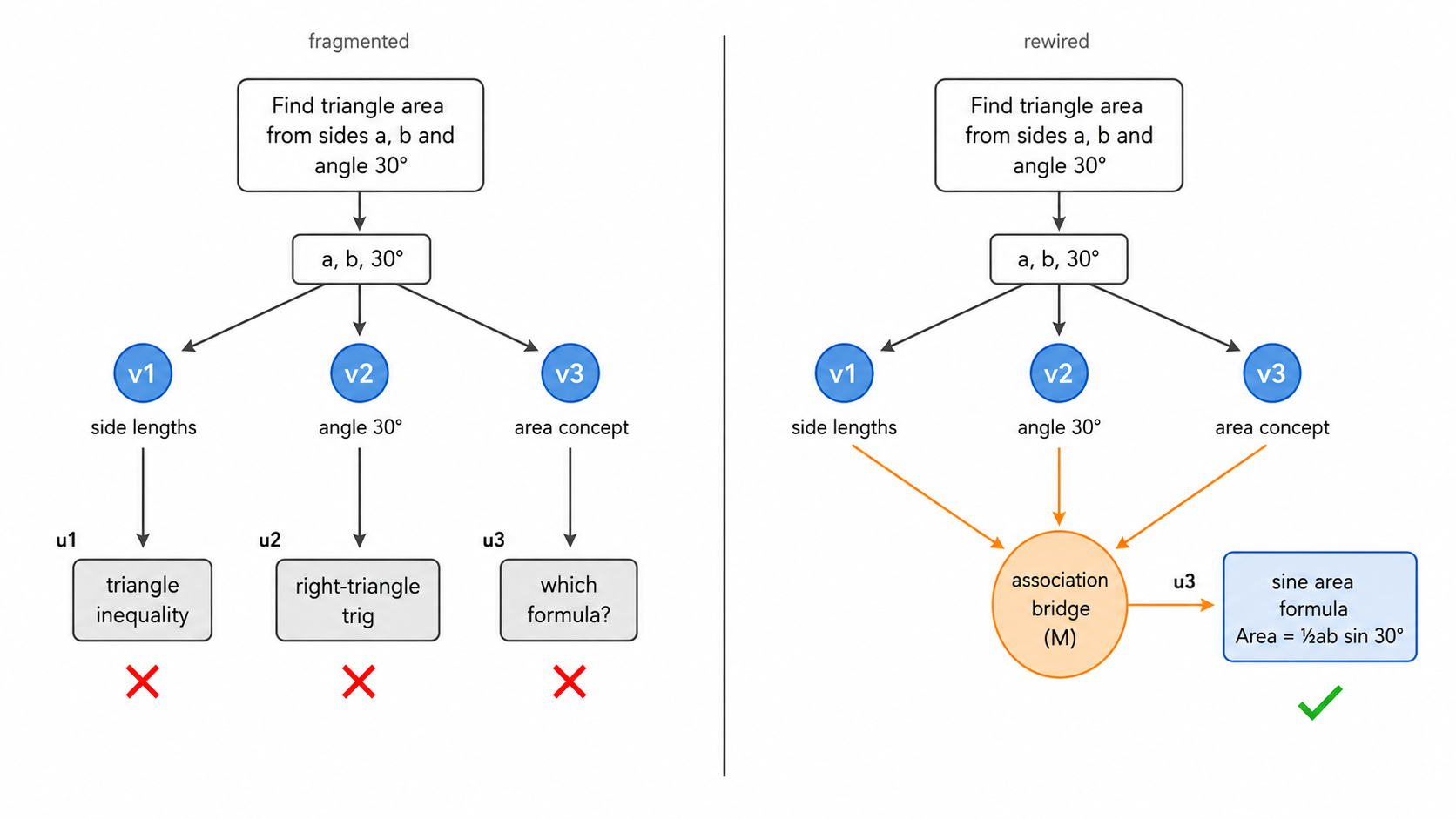}
    \caption{Illustration of spectral rewiring. The base model primarily performs point-to-point associations between isolated spectral components. RL-induced rewiring introduces many-to-one routing, allowing multiple premise directions to jointly activate an output direction associated with a compositional reasoning step.}
    \label{fig:spectral_rewiring_example}
\end{figure}

\section{Implementation Details}
\label{app:evaluation_config}

\paragraph{Algorithmic Scope.}
For each Transformer block, SAR is applied only to the attention linear weights ($W_q,W_k,W_v,W_o$) and MLP linear weights ($W_{\mathrm{gate}},W_{\mathrm{up}},W_{\mathrm{down}}$). We do not modify biases, normalization parameters, token embeddings, or the language-modeling head; these parameters are kept from the base model. Empirically, projecting both attention and MLP weights is necessary to recover the full effect: projecting only attention or only MLP degrades relative to the full-RL model, while using the base or RL versions of the embeddings and lm\_head has little effect on the final results.

\paragraph{Evaluation Configuration.} Unless otherwise specified, we use the same decoding configuration for all reasoning and coding evaluations. We sample with temperature $T=0.6$, top-$p=0.95$, and a maximum generation length of $32{,}768$ tokens. For POLARIS, we follow its longer-generation setting and use a maximum generation length of $96{,}000$ tokens. All compared models within the same experiment use identical sampling parameters to ensure a fair comparison between the base model, full RL model, SAR-projected model, and no-projection control.

\paragraph{Numerical Precision.}
All SVD decompositions and SAR projections are computed in FP32. Since the reconstructed SAR update is compact and sensitive to rounding, directly storing the projected weights in low precision can perturb the intended spectral geometry, especially for smaller models. We therefore store and evaluate the 1.5B, 4B, and 7B models in FP16, which better preserves fine-grained updates than BF16. For the 32B OLMo-3 models, we store the projected weights in FP16 and evaluate with BF16 computation, as they are empirically less sensitive to floating-point perturbations. Unless otherwise stated, all compared models within the same experiment use the same runtime precision.

\paragraph{Coverage analysis.}
We report a coverage-based view of the large-$k$ improvement in Table~\ref{tab:coverage_256}. A problem is counted as covered if any of 256 rollouts produces the correct final answer. Under this metric, SAR solves one additional AIME 2024 problem beyond the full RL model, indicating that the projected model expands the set of reachable correct solutions rather than merely changing single-sample accuracy.

\begin{table}[t]
\centering
\caption{Coverage analysis on AIME 2024 with 256 rollouts. A problem is counted as covered if any rollout produces the correct final answer.}
\label{tab:coverage_256}
\begin{tabular}{lccc}
\toprule
Model & Coverage@256 & Solved / Total & Gain \\
\midrule
DeepScaleR full RL & 83.33\% & 25 / 30 & -- \\
SAR, 1\% rank & 86.66\% & 26 / 30 & +1 problem \\
\bottomrule
\end{tabular}
\end{table}

\section{Boundary Cases of Spectral Rewiring}
\label{app:boundary_cases}

SAR is most effective when RL remains in an elicitation regime, where sparse verifiable rewards and token-averaged updates produce relatively small per-step changes that reorganize capabilities already encoded in the reference model. We observe several boundary cases where this assumption does not fully hold.

\paragraph{Heavily trained reasoning RL.}
For reasoning models trained with many RL steps, the accumulated update can move beyond compact spectral elicitation. In our JustRL case study, which is trained for over 4000 steps \cite{justrl2026scaling}, projecting the RL update into the base spectral space recovers a non-trivial AIME 2024 score of roughly 43\%. This is comparable to the earlier DeepScaleR-level result, but remains below the final RL model ($\sim 52\%$). The result suggests that the base spectral manifold still contains meaningful recoverable reasoning capacity, while the remaining gain may come from stronger task-specific parameter rewriting induced by extended optimization. We therefore interpret this case as a boundary between elicitation and rewriting, rather than as evidence that spectral structure is absent.

\paragraph{Direct code RL from a base model.}
Code RL introduces domain requirements beyond abstract reasoning. The model must satisfy an execution protocol: syntactically valid code, function-completion or stdin/stdout conventions, exact input-output formatting, unit-test behavior, and edge-case handling. These behaviors can require adapting programming knowledge and interface behavior that are not sufficiently established in the base model's spectral geometry. Consistent with this interpretation, direct code RL from the base model is less projection-compatible in our experiments, as in the DeepCoder case \cite{together2025deepcoder}. However, once code SFT establishes the relevant programming interface, the subsequent RL update becomes more compatible with spectral projection onto the SFT model's spectral space. For example, projecting X-coder-RL onto X-coder-SFT recovers the full RL performance \cite{wu2026xcoder}. Our OLMo-3 experiments further support this view: after coding-oriented training establishes the relevant domain knowledge, RL primarily elicits reasoning capability already present in the reference model.

\paragraph{Dense critic rewards and PPO-style training.}
PPO-style training with an external critic or dense reward can also reduce projection compatibility. The key factor is not the PPO algorithm itself, but the effective optimization strength: dense token-level rewards and critic-shaped objectives can produce larger and more continuous parameter movement than sparse outcome-reward RL. This can push the update away from compact rewiring in the reference spectral coordinates and toward broader behavioral rewriting. Thus, SAR should be understood as most directly characterizing sparse outcome-reward elicitation, while dense-reward RL may require either a different reference point, intermediate projection during training, or a richer spectral basis.

Together, these cases delineate the scope of SAR. The method captures the component of RL that is aligned with the reference model's spectral manifold. When the reference model already contains the relevant reasoning ability or domain knowledge, this component can account for the dominant effect of RL. When training intensity or missing domain knowledge forces the model to acquire behavior outside that manifold, spectral projection remains informative but becomes an incomplete reconstruction of the full RL update.

\section{Comparison with the No-Projection Control}
\label{app:no_projection_control}

To isolate the effect of spectral alignment, we compare SAR with a no-projection control. This control extracts the same top-ranked low-rank component from the RL update $\Delta W$, but directly adds it back to the base model without projecting it onto the pretrained SVD subspace. Thus, both methods use the same rank budget, and the only difference is whether the compact update is aligned with the base model's spectral geometry.

\begin{figure}
    \centering
    \includegraphics[width=0.8\linewidth]{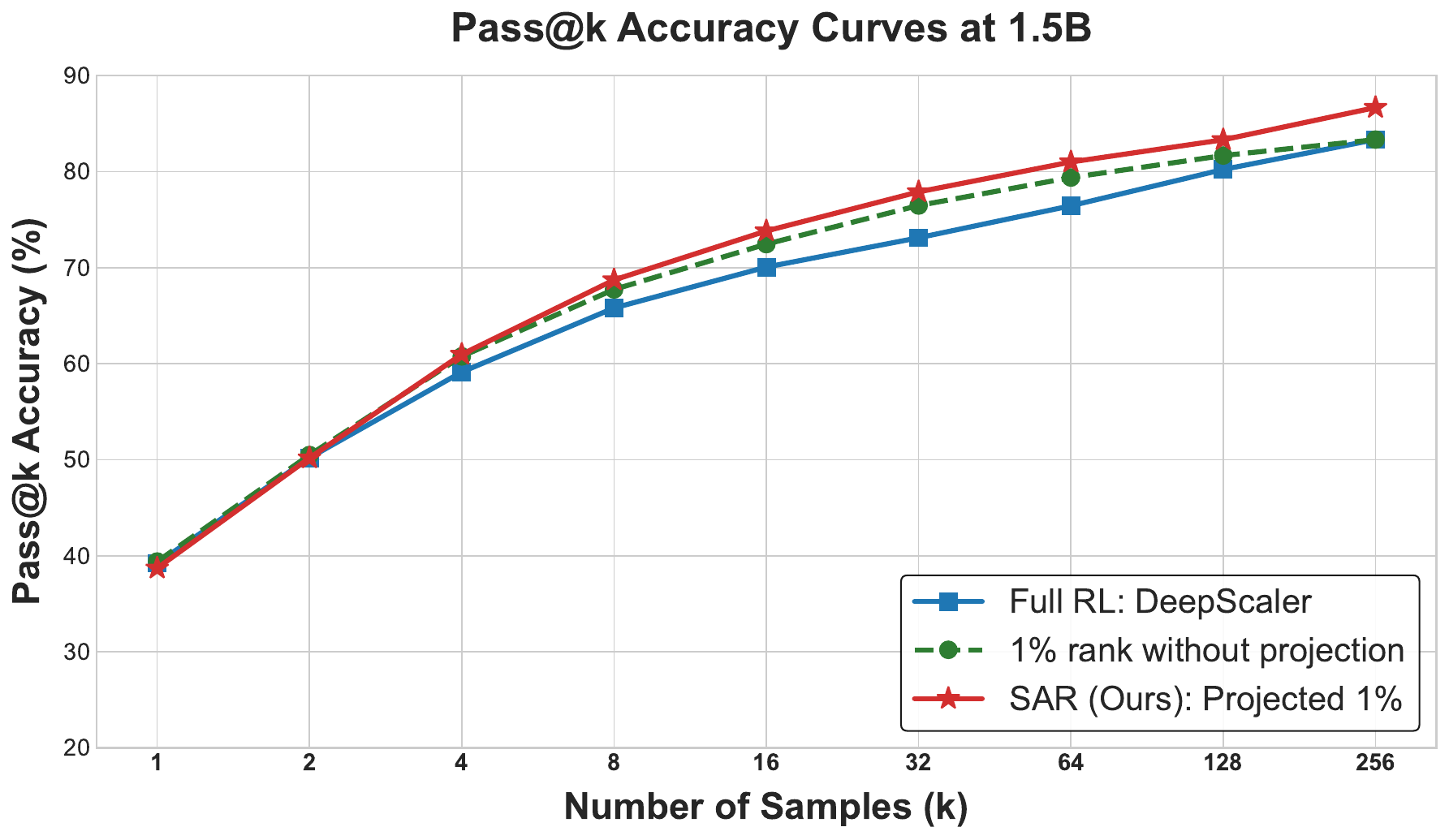}
    \caption{Pass@$k$ scaling on AIME 24 among full RL, SAR, and the low-rank component without projection.}
    \label{fig:no_projection_1.5B aime 24}
\end{figure}

\paragraph{Reasoning and Exploration.}
We first evaluate reasoning and exploration on AIME 2024 using the DeepScaleR models. For each problem, we generate 256 rollouts and report the Pass@$k$ scaling behavior in Figure~\ref{fig:no_projection_1.5B aime 24}. SAR and the no-projection control achieve similar Pass@1 performance, showing that low-rank extraction alone can preserve the single-sample reasoning gains of RL. However, their large-$k$ behavior differs clearly: the no-projection control saturates earlier, while SAR continues to improve. At Pass@256, SAR solves one additional problem that neither the full RL model nor the no-projection control solves within 256 attempts. This indicates that the reasoning-effective update is not merely low-rank; aligning it with the base model's SVD space is \textbf{important for recovering reasoning and exploration capability}.

\paragraph{Multi-Domain RL.}
We then evaluate the same control in the multi-domain setting using the OLMo-3-32B models on AIME 2025 and LiveCodeBench v5. Results are shown in Table~\ref{tab:aime_lcb_noproj}. Consistent with the single-domain results, the no-projection control retains part of the AIME performance, but fails to recover SAR's exploration benefit. On AIME 2025, the control reaches only 86.67\% Pass@32, whereas SAR reaches 93.33\%. On LiveCodeBench, the control recovers part of the performance, suggesting that low-rank extraction can remove some redundant update directions. Nevertheless, SAR achieves the strongest coding performance, improving AVG@10 to 69.09\%, compared with 68.24\% for the control and 67.61\% for full RL.

\paragraph{Cross-Domain Merging.}
We also evaluate a no-projection control for math-to-code expert merging. This control extracts the same 1\% low-rank component from the math RL update and merges it into the code expert, but does not align the update with the pretrained spectral basis. As shown in Table~\ref{tab:merge_noproj_control}, the no-projection merge preserves the code-side gain, but substantially weakens math reasoning compared with SAR. This indicates that low-rank extraction alone is insufficient for positive-sum composition; the spectral projection step is what makes the math reasoning update compatible with the code expert.

Overall, this comparison supports our central insight: the reasoning-effective parameters are not only compact, but are specifically \textbf{expressed in the base model's pretrained SVD space}. Low-rank extraction explains part of the parameter efficiency, but SAR's spectral alignment is what yields stronger exploration and cross-domain transfer. In several settings, SAR even outperforms the full RL model, indicating that extracting the compact spectral reasoning core is not merely a compression technique, but a useful model-editing operation in its own right.

\begin{table}[t]
\centering
\caption{Head-to-head comparison on AIME 2025 and LiveCodeBench v5. The no-projection control keeps the same top-1\% low-rank component but removes the projection onto the pretrained SVD subspace. SAR achieves the strongest large-$k$ reasoning coverage and coding performance.}
\label{tab:aime_lcb_noproj}
\resizebox{\textwidth}{!}{
\begin{tabular}{lcccc}
\toprule
\multirow{2}{*}{\textbf{Model / Method}} 
& \multicolumn{2}{c}{\textbf{AIME 2025}} 
& \multicolumn{2}{c}{\textbf{LiveCodeBench v5}} \\
\cmidrule(lr){2-3} \cmidrule(lr){4-5}
& \textbf{AVG@32} & \textbf{Pass@32} & \textbf{AVG@10} & \textbf{Pass@10} \\
\midrule
Base: OLMo3-32B-Think-DPO 
& 70.83\% & 90.00\% & 64.81\% & 78.07\% \\
Full RL: OLMo-3.1-32B-Think 
& \textbf{75.76\%} & 90.00\% & 67.61\% & 80.91\% \\
No projection, 1\% 
& 75.50\% & 86.67\% & 68.24\% & 81.13\% \\
\textbf{SAR projection}, 1\% 
& 75.16\% & \textbf{93.33\%} & \textbf{69.09\%} & \textbf{81.59\%} \\
\bottomrule
\end{tabular}
}
\end{table}

\begin{table*}[t]
\centering
\renewcommand{\arraystretch}{1.1}
\caption{No-projection control for 1.5B math-code merging. The no-projection merge uses the same 1\% low-rank math update budget as SAR but skips alignment to the pretrained spectral basis. It preserves code performance but loses much of the math gain, whereas SAR achieves positive-sum transfer.}
\label{tab:merge_noproj_control}
\resizebox{0.8\textwidth}{!}{
\begin{tabular}{llccc}
\toprule
\multirow{2}{*}{\textbf{Scale}} & \multirow{2}{*}{\textbf{Model / Method}} & \multicolumn{2}{c}{\textbf{Math (AIME 24)}} & \textbf{Code (LCB)} \\
\cmidrule(lr){3-4} \cmidrule(lr){5-5}
& & \textbf{AVG@32} & \textbf{Pass@32} & \textbf{AVG@8} \\
\midrule

\multirow{8}{*}{\textbf{1.5B}} 
& \multicolumn{4}{l}{\textit{\color{gray}Single-Domain References}} \\
& Base: Distill-Qwen-1.5B & 31.67\% & 76.67\% & 23.00\% \\
& Math Expert: DeepScaleR & \textbf{40.31\%} & \textbf{76.67\%} & 27.20\% \\
& Code Expert: Archer-Code & 39.48\% & 76.67\% & \underline{31.80\%} \\
\cmidrule(lr){2-5}
& \textit{Best Single-Domain Expert} & \textit{40.31\%} & \textit{76.67\%} & \textit{31.80\%} \\
\cmidrule(lr){2-5}
& \multicolumn{4}{l}{\textit{\color{gray}Cross-Domain Merging}} \\
& No Projection: 1\% Math-to-Code & 36.25\% & 63.33\% & \underline{32.25\%} \\
& \cellcolor{gray!15}\textbf{Spectral Projection Merge (SAR)} & \cellcolor{gray!15}\textbf{43.44\%} & \cellcolor{gray!15}\textbf{76.67\%} & \cellcolor{gray!15}\textbf{32.25\%} \\
\bottomrule
\end{tabular}
}
\end{table*}

\section{Method Ablations}
\label{app:method_ablations}

We further ablate the structure of SAR on DeepScaleR-1.5B using the same 1\% spectral budget on AIME 2024. Table~\ref{tab:method_ablation} compares SAR with three controls: replacing the pretrained spectral basis with a random projection, retaining only the diagonal entries of the rewiring matrix $M$, and retaining only its off-diagonal entries.

\begin{table}[t]
\centering
\caption{Method ablations on DeepScaleR-1.5B with a 1\% update budget on AIME 2024. SAR preserves full-RL-level Pass@1 while maintaining Pass@32, whereas random projection and partial rewiring controls lose reliability or coverage.}
\label{tab:method_ablation}
\begin{tabular}{lcc}
\toprule
\textbf{Method} & \textbf{Pass@1} & \textbf{Pass@32} \\
\midrule
Base & 31.67\% & 80.00\% \\
Full RL & \textbf{40.31\%} & 76.67\% \\
SAR, 1\% & 40.21\% & 76.67\% \\
\midrule
Random projection, 1\% & 36.67\% & 80.00\% \\
Diagonal-only rewiring, 1\% & 30.73\% & 80.00\% \\
Off-diagonal-only rewiring, 1\% & 38.54\% & 73.33\% \\
\bottomrule
\end{tabular}
\end{table}

These ablations support three conclusions. First, the random projection baseline improves over the base model in Pass@1 but remains clearly below SAR, showing that the pretrained singular-vector coordinates are important rather than interchangeable with an arbitrary low-dimensional projection. Second, the diagonal-only variant fails to recover the RL gain, suggesting that simply rescaling isolated pretrained components is insufficient for reasoning elicitation. Third, the off-diagonal-only variant recovers most of the Pass@1 gain but loses Pass@32 coverage, indicating that cross-component rewiring carries a large part of the reasoning signal but works best together with the diagonal component. Overall, SAR's effect comes from the combination of pretrained spectral alignment and the full rewiring matrix, rather than from low-rank structure, random projection, or either matrix component alone.

\end{document}